\documentclass[lettersize,journal]{IEEEtran}
\usepackage{amsmath,amsfonts}
\usepackage{array}
\usepackage{algorithmicx,algorithm}
\usepackage[caption=false,font=normalsize,labelfont=sf,textfont=sf]{subfig}
\usepackage[noend]{algpseudocode}
\usepackage{textcomp}
\usepackage{stfloats}
\usepackage{url}
\usepackage{verbatim}
\usepackage{graphicx}
\usepackage{cite}
\usepackage{hyperref}
\usepackage{booktabs}
\usepackage{blindtext}
\usepackage{tabularx}
\usepackage{multirow}
\usepackage{pifont}
\usepackage[capitalize]{cleveref}
\usepackage{bbold}
\usepackage{amssymb}
\usepackage{indentfirst}
\usepackage{autobreak}
\usepackage{tikz}
\usepackage{epstopdf}
\usepackage{diagbox}
\usepackage{epstopdf}
\usepackage{wasysym}
\usepackage{balance}
\usepackage{makecell}
\usepackage[normalem]{ulem}
\crefname{section}{Sec.}{Secs.}
\Crefname{section}{Section}{Sections}
\Crefname{table}{Table}{Tables}
\crefname{table}{Tab.}{Tabs.}
\hypersetup{hypertex=true,
            colorlinks=true,
            linkcolor=blue,
            anchorcolor=blue,
            citecolor=blue}

\begin{document}

\title{Prototypical Progressive Alignment and Reweighting for Generalizable Semantic Segmentation}

\author{Yuhang Zhang, Zhengyu Zhang, Muxin Liao, Shishun Tian, Wenbin Zou$^*$, Lu Zhang, Chen Xu
        % <-this % stops a space
\thanks{Yuhang Zhang is with the School of Computer Science and Cyber Engineering, Guangzhou University, Guangzhou, China (e-mail: hectorcheung@foxmail.com).}
\thanks{M. Liao, S. Tian, are with Guangdong Key Laboratory of Intelligent Information Processing, College of Electronics and Information Engineering, Shenzhen University, Shenzhen, 518060, China (e-mail: liaomuxin2020@email.szu.edu.cn, stian@szu.edu.cn).}
\thanks{Z. Zhang is with the School of Electronics and Communication Engineering, Guangzhou University, Guangzhou, China (e-mail: zhengyuzhang@gzhu.edu.cn).}
\thanks{L. Zhang are with the Univ Rennes, INSA Rennes, CNRS, IETR - UMR 6164, F-35000 Rennes, France (e-mail: lu.ge@insa-rennes.fr).}
\thanks{W. Zou is with Guangdong Key Laboratory of Intelligent Information Processing, Shenzhen Key Laboratory of Advanced Machine Learning and Applications, Institute of Artificial Intelligence and Advanced Communication, College of Electronics and Information Engineering, Shenzhen University, Shenzhen, 518060, China (e-mail: wzou@szu.edu.cn). (Corresponding author: Wenbin Zou.).}
\thanks{C. Xu is with the College of Mathematics and Statistics, Shenzhen University, Shenzhen 518060, China (e-mail: xuchen@szu.edu.cn).}}

\markboth{Journal of \LaTeX\ Class Files,~Vol.~14, No.~8, August~2021}%
{Shell \MakeLowercase{\textit{et al.}}: A Sample Article Using IEEEtran.cls for IEEE Journals}


\maketitle

\begin{abstract}
Generalizable semantic segmentation, aims to excel on unseen target domains, as a critical focus due to the widespread practical applications requiring high generalizability. Class-wise prototypes, which depict class-wise centroids, as a type of domain-invariant information are key to improving the model generalizability due to its stability and representativeness. However, this manner faces some challenges. First, the existing methods adopt a coarse prototypical alignment form, potentially compromising performance. Second, the naive prototype generally serves as the class centroid generated by an average operation from source data batches, risks source domain overfitting, and may be detrimentally impacted by unrelated source data. Third, from a broader perspective, rather than just from a prototypical alignment perspective, the existing methods treat all samples equally, which is against the conclusion that different source features have different adaptation difficulties. To tackle these issues, we propose a novel method for generalizable semantic segmentation called Prototypical Progressive Alignment and Reweighting (PPAR) depending on the strong generalized representation of the Contrastive Language-Image Pretraining (CLIP) model. In particular, we first define the Original Text Prototype (OTP) and Visual Text Prototype (VTP) generated by the CLIP model, laying the foundation for the subsequent effective alignment strategy. Then, we propose a prototypical progressive alignment strategy by an easy-to-difficult alignment form to reduce domain-variant information progressively instead of directly. Finally, we propose a prototypical reweighting learning strategy that estimates the importance of the source data and corrects its learning weight to alleviate the influence of unrelated source features, i.e. alleviate negative transfer. Moreover, we also offer a theoretical insight into our method and it shows that our method compiles well on the domain generalization theory. Extensive experiments on several popular datasets demonstrate that our PPAR method achieves superior performance, proving the effectiveness of our method.

\end{abstract}


\section{Introduction}
\par Intelligent Transportation Systems (ITSs) is a comprehensive field focusing on improving transportation effectiveness, widely-used in automatic driving, moving robotics, and so on. These ITSs heavily rely on perception tasks, which are vital for interpreting the surrounding environment and making informed decisions. Among perception tasks, semantic segmentation \cite{yan2022threshold} plays a crucial role in accurately identifying objects such as vehicles, pedestrians, and traffic signs, ensuring the system can operate safely and effectively, which aims to give the correct category label to each pixel of images \cite{wang2022sfnet, hua2023multiple}. Recently, several types of popular paradigms have been made by researchers. As a typical manner, supervised semantic segmentation \cite{hu2024contrastive} has achieved superior performance without considering domain divergence between the training and validation sets, which is kindly applied in unchanged environments while suffering a performance drop in changed environments due to the domain shift problem \cite{yang2023improving} (i.e., there is a domain divergence between the training and validation sets).

\par Limited by the massive cost of data labeling \cite{ru2023token} and the inability to tackle the domain shift problem \cite{liao2022exploring}, unsupervised domain adaptation (UDA) was developed as a new task \cite{zhang2021transfer, zou2023dual}, which aims to achieve remarkable performance in the target domain given the labeled source data and unlabeled target data. The domain gap, which harms the performance in the target domain, is reduced by some efficient UDA strategies such as distribution alignment \cite{zou2023dual} and image style transfer \cite{yang2020fda}. Nevertheless, the available target data is a primary obstruction in practical applications since many applications have privacy protection requirements. Moreover, performing well in one fixed target environment can not satisfy practical requirements owing to the diversity in the real world.
\begin{figure}[t]
	\centering
	\includegraphics[width=0.48\textwidth]{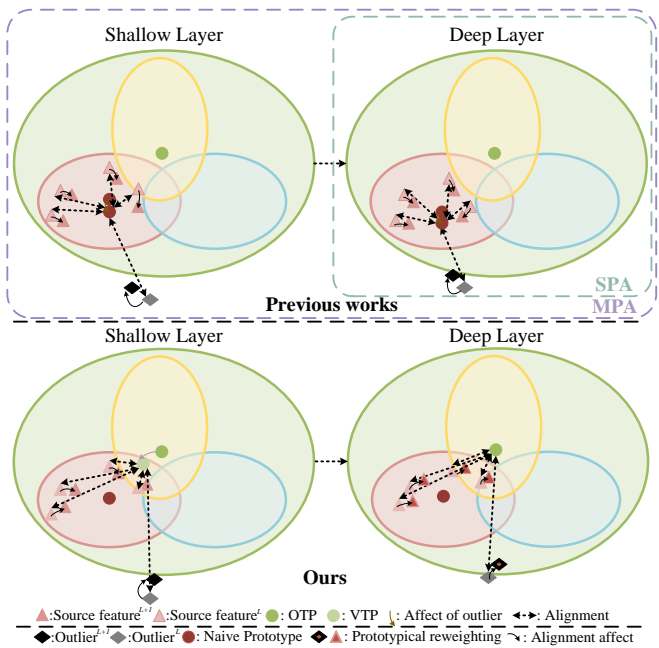}
	\caption{Approach comparison. The red ellipsoid denotes the source domain, the green one represents the overarching unknown target domain, while the yellow and blue ellipsoids indicate specific target domains. In prototypical reweighting, the red percent in features represents the learning weight. The source outlier has a relatively low learning weight.}
	\label{Fig:PROTOCOMP}
\end{figure}

\par More recently, domain generalization (DG) has attracted a lot of attention to tackle these above drawbacks in UDA, with the goal of generalizing well on all probable unseen domains \cite{wang2022feature}. Different from the UDA, the target domain data do not participate in the training period of DG and is only used for evaluation \cite{wang2022generalizing}. Designing a DG model with high generalizability generally obeys a core rule that extracts domain-invariant features, which can be roughly devised from two aspects, i.e., style extension and style erasion. Style extension \cite{yue2019domain, huang2021fsdr} generates images with different styles as the extended training set to enhance the data diversity. Style erasion uses some representation constraints like normalization \cite{ioffe2015batch}, whitening \cite{pan2019switchable}, and prototypical alignment \cite{liu2021bapa}. Among them, prototypical alignment is an effective method for domain adaptation and domain generalization, which pulls close the class-wise features to the related class prototype (i.e., class centroid) \cite{zhang2019category}. In the prototypical alignment paradigm, the class-wise prototype is expected as the class-wise domain-invariant feature since that can be seen as the combined representation of multi-domains. To learn domain-invariant features from the class prototype, three key observations emerge as follows.
\begin{enumerate}
\item The naive prototype can be regarded as the source-bias prototype that provides a weak generalizable representation for prototypical alignment, as typical approaches \cite{liu2021bapa, jiang2022prototypical, lee2022bi} generate the prototype by computing the class centroid from a batch of source data rather than using overall domain data. The overall domain denotes the potential domain consisting of the source and many unseen sub-target domains, which is represented by the green ellipsoid in Fig. \ref{Fig:PROTOCOMP}. It can be seen that the naive prototype (i.e., red circle) is generated by several source samples, which can only reflect the source centroid, rather than the overall centroid (i.e., green circle) that contains more generalizable information. Aligning features with such naive prototypes may drive the model towards source-biased prototypes, increasing the risk of overfitting to the source domain and susceptibility to source outliers.
\item The existing prototypical alignment methods can be broadly categorized into single-layer prototypical alignment (SPA) and multi-layer prototypical alignment (MPA), as illustrated in Fig. \ref{Fig:PROTOCOMP}. SPA performs alignment only at a specific layer, typically the final one \cite{liu2021bapa}, focusing on a single-layer representation. However, this approach results in insufficient alignment, leading to suboptimal performance, as it fails to consider the multi-scale alignment of features. In contrast, MPA attempts to align features across multiple layers, capturing deeper information but often neglecting the unique attributes of shallow features \cite{jiang2022prototypical}, such as their ability to capture basic visual information or style cues \cite{gatys2016image, zhang2018fully}.  Although multi-scale alignment methods have demonstrated better performance \cite{huang2020contextual, guan2021scale}, MPA's uniform treatment of shallow and deep features can inadvertently remove useful information from the shallow layers, which play a crucial role in encoding fundamental visual details. An refined approach is required to fully leverage both shallow and deep features without discarding valuable low-level information, thus addressing the limitations of both SPA and MPA. The rectangle with rounded corners and dashed edges of Fig. \ref{Fig:PROTOCOMP} provides an intuitive visual abstract for these two strategies. 
\item From a broader perspective, rather than just from a prototypical alignment perspective, the existing methods treat all samples equally, which is against the conclusion that different source features have different adaptation difficulties  \cite{yu2021dast, wang2019characterizing}. This means that the model cannot determine the importance of features, thereby forcibly learning knowledge from the unrelated source feature. 
\end{enumerate}
\begin{comment}

\textcolor{blue}{Prototype alignment-based methods stand on prototype construction and alignment form. There are some shortcomings in these two sub-parts. For prototype construction, previous methods generate prototypes by averaging the source feature, i.e., calculating class centroids. As BCL\cite{lee2022bi} mentioned that there is a discrepancy between the class centroids of the source and target domains, thus such prototype construction may lead to overfitting to the source domain and may suffer from the influence of unrelated source data. For alignment form, some methods only focused on the prototypical distribution alignment in features of the last layer \cite{liu2021bapa}, i.e., single-layer prototypical alignment (SPA). Another method \cite{jiang2022prototypical} holds that multi-layer prototypical distribution alignment (MPA) plays a more effective role in domain-invariant feature extraction. The former does not constrain the shallow features, which may lead to suboptimal performance. Although the latter constrains the shallow feature, the prototypical alignment in the shallow feature adopts the same way as that in the deep feature, which may remove useful basic visual information in the shallow features.}
\end{comment}
\par Enlightened by the above analyses, in order to perform prototypical learning in a refined manner, we develop a novel framework for generalized semantic segmentation called Prototypical Progressive Alignment and Reweighting (PPAR) depending on the strong generalized representation provided by the Contrastive Language-Image Pretraining (CLIP) model \cite{radford2021learning}. First, for the prototype construction, we defined the original text prototype (OTP) and visual text prototype (VTP) generated by the CLIP model to replace the naive prototype, which serves as the foundation for the subsequent efficient and effective alignment strategy. Second, for the alignment form, inspired by some curriculum learning UDA methods \cite{zhang2017curriculum, sakaridis2020map} that gradually perform domain adaptation for better knowledge transfer, while fully considering the attributes of shallow and deep features, we propose a prototypical progressive alignment strategy adopting an easy-to-difficult alignment form that achieves better generalizability of the model. Moreover, to reduce the influence of the outlier features causing the negative transfer, a prototypical reweighting learning strategy is proposed, which targets estimating the importance of the source data and correcting its learning weight.
\par We also depict the core of our method in Fig. \ref{Fig:PROTOCOMP}. Compared to the naive prototype, the proposed prototype (i.e., light and dark green circle) is closer to the overall domain since the CLIP model learns knowledge from large-scale and multiple datasets, consequently aligning features (i.e., triangle) more effectively toward a more generalized space. For progressive alignment, from a local perspective, since the VTP contains more basic visual information about the source domain, VTP is viewed as the intermediate domain (called the source-overall intermediate domain) between the source and the overall domains, serving as a bridge role. Therefore, shallow alignment is a more effective distribution alignment between the source and the source-overall intermediate domains since their domain gap is smaller than that between the source and overall domains. This viewpoint is evidenced in some previous works \cite{xu2020adversarial, wu2020dual, zhang2023hybrid}. From a global perspective, this strategy can be regarded as a curriculum learning strategy, i.e., the learning domain-invariant knowledge from easy to difficult (from VTP to OTP). From the learning perspective, the feature far from the OTP can be regarded as the source outlier is given a low learning weight.

\par In a nutshell, the main contributions of this paper can be concluded as follows.
\begin{itemize}
\item We propose a method for generalizable semantic segmentation called Prototypical Progressive Alignment and Reweighting (PPAR) depending on the strong generalized representation of the CLIP model, which serves as a pioneering template in this field for integrating the large model with the downstream task.

\item To further improve the effectiveness of prototypical alignment, we propose a prototypical progressive alignment strategy in an easy-to-difficult alignment form, progressively reducing the domain divergence.

\item To alleviate the influence of the outlier feature and reduce negative transfer, we propose a prototypical reweighting learning strategy, which estimates the importance of the source data and corrects its learning weight.

\item Our PPAR approach achieves superior performance on several challenging benchmarks and outperforms other state-of-the-art methods.
\end{itemize}

\section{Related work}

%-------------------------------------------------------------------------
\subsection{Semantic segmentation}
Semantic segmentation, as known as scene understanding, is widely used in autonomous driving, which assigns different colors to each pixel to identify diverse classes \cite{pan2022deep, zhang2023delivering}. Typically, three mainstream segmentation architectures can be representatively shown by fully convolution networks (FCNs) \cite{long2015fully}, SegNet \cite{badrinarayanan2017segnet}, and DeepLab-series networks \cite{chen2017deeplab, chen2017rethinking, chen2018encoder}, respectively, where SegNet adopts a standard encoder-decoder mechanism and DeepLab-series networks deeply verified the advantage of multi-scale atrous convolution. From this consideration, many complex segmentation architectures have emerged, such as DenseNet \cite{huang2017densely}, HRNet \cite{sun2019deep, yu2021lite}, SegNeXt\cite{guosegnext}, and Transformer form like Segmenter \cite{strudel2021segmenter} and MHVT \cite{gu2022multi}, which fully connect, fuse, and recover the lost semantic information due to down-sampling operation. With the inspiration of human vision that focuses on the most salient part, some segmentation frameworks presented an attention-based paradigm, such as GCNet \cite{cao2019gcnet}, CCNet \cite{huang2019ccnet}, and PIDNet \cite{xu2023pidnet}, aiming to capture the most salient region of semantics for better discrimination. Some approaches  \cite{romera2017erfnet, gao2021mscfnet} developed lightweight models for easily deploying to the hardware. To perceive the 3D structure of the object, some approaches introduce LiDAR \cite{chang2023multi} and depth sensors \cite{shi2021rgb} to obtain geometry information. Although these methods elicited impressive results, they depended on large-scale data including pixel-level annotations with enormous consumption. Meanwhile, they suffer from the domain shift problem, that is, performance decreases when the environment changes.

\subsection{Unsupervised domain adaptation}
Domain adaptation transfers knowledge from the source domain to the target domain \cite{zhang2022confidence}. According to the annotation of the target domain, domain adaptation can be parted into a weak-supervised or unsupervised manner. Most methods focus on unsupervised domain adaptation (UDA) since the UDA task only uses the unlabeled target data, which fully minimizes the label collecting \cite{yan2022threshold}. The existing UDA methods can be roughly divided into image translation, representation distribution alignment, self-training, and domain mixing. The image translation methods focus on transferring the source style to the target style, which reduces the domain gap from the image level \cite{yang2020fda, kim2020learning}. The representation distribution alignment methods try to perform distribution alignment at the feature level such as adversarial learning \cite{zhou2020affinity, zou2023dual}, and prototypical alignment \cite{liu2021bapa, jiang2022prototypical, lee2022bi}. The self-training methods fine-tune the segmentation model using the target image and related pseudo label, which mainly focuses on the design of high-quality pseudo label generation \cite{zou2019confidence, zheng2021rectifying}. Meanwhile, the self-training methods are verified that could be combined with other UDA methods further improve the performance in the target environment \cite{yu2021dast, liao2022exploring}.  In the context of these strategies, the work of Luo et al. \cite{luo2022towards} stands out as it cleverly combines meta-learning and mixed sampling strategy to address the multi-source UDA problem. Domain mixing strategy, another image-level UDA method, uses two domain data to generate mixed images, which turns the unsupervised problem into a semi-supervised problem \cite{tranheden2021dacs, zhou2022context}. Recently, some works verified that transformer-based architecture is beneficial for the UDA task \cite{hoyer2022daformer}. Although these UDA methods achieve remarkable performance, these methods can not generalize well on other unseen domains. Furthermore, the target domain data of UDA is available at training, which can not be satisfied in many practical applications.

\subsection{Domain generalization}
Domain generalization (DG) aims to achieve superior performance on all unseen domains  \cite{liao2023domain}. Instead of training with the source and target domains, the DG model is trained with only the source domain. According to the number of source domains, DG settings are parted into single-domain generalization and multi-domain generalization. The single-domain generalization is more challenging since only one source domain can be used in the training period, which is the goal of this paper. Data augmentation, representation learning, and meta-learning are the three main DG approaches. Data augmentation performs domain randomization (i.e., a transformation for the input data) at the image level or feature level to cover the styles of probable unseen domains. DRPC \cite{yue2019domain} mapped the style of auxiliary domains into the source domains and aligned the different results from the same model. GTR \cite{peng2021global} considered global and local domain randomization using a painting auxiliary dataset to refine the distribution alignment. SHADE \cite{zhao2022style} randomized the style of the data by only the source domain without extra data. Representation learning normalizes or whitens the deep features to reduce feature discrepancy, which plays a style-reduction role \cite{pan2018two, pan2019switchable,choi2021robustnet, peng2022semantic}. Meta-learning is a novel paradigm that emerged in recent works to simulate the domain shift based on a learning-to-learn mechanism \cite{zhang2022generalizable, kim2022pin}. FGSS \cite{zhang2023fine} presented a fine-grained self-supervision framework from the perspective of the intra-class relationship, which forged a new path for generalizable semantic segmentation. Recently, CSDS \cite{liao2024class} focuses on rare classes sample and feature alignment. CMPCL \cite{liao2024calibration} proposed a uncertainty-weighted and hard-weighted multi-prototype contrastive learning strategy.
\subsection{CLIP-based semantic segmentation}
Owing to the development of multi-modal and the importance of text information, CLIP is also used in some fields, such as referring semantic segmentation and open-vocabulary semantic segmentation. For the former, this task aims to segment corresponding objects in an image by giving the text string. CRIS \cite{wang2022cris} designed a long-range vision-language decoding module and contrastive learning for text-to-pixel alignment. TRIS \cite{liu2023referring} proposed a weakly-supervised referring semantic segmentation framework owing to the sufficient localization ability of the text information, by only using text supervision. UDA-RSS \cite{shi2023unsupervised}, with the aims of the unsupervised domain adaptation in the referring semantic segmentation, proposed to generate pseudo-text for the target domain and then perform distribution alignment. For the latter, open vocabulary/zero-set semantic segmentation targets to classify arbitrary and unseen categories. ZSSeg \cite{xu2022simple} proposed a two-stage framework to achieve this, which fixed feature generated by the CLIP text extractor as the classifier weight to learn a zero-shot image encoder, i.e., maximum the similarity between image feature and text feature under the same class. As this manner needs	 expensive annotations, some methods like GroupViT \cite{xu2022groupvit} and ViL-Seg \cite{liu2022open} achieved this only using text supervision, with the core of contrast learning between text and image features. In summary, these methods achieved better generalizability to unseen classes by capturing the relationship between image-text features and between different classes.

\section{Preliminaries}
\label{Sec:Pre}
Prototypical alignment, as a well-known strategy, is widely used in many UDA methods  \cite{liu2021bapa, jiang2022prototypical, lee2022bi}, which effectively reduces the domain gap. The prototype is generally explained as the class centroid, which is a generalized representation at the class level. Given a batch of source domain data including the images and related annotations $\{x_s\in X_s, y_s \in Y_s\}$, the feature $f^n$ of class $n$ in (h,w) spatial location can be extracted as:
\begin{equation}
{f^n}^{(h,w)} =F(x_s)^{(h,w)}\mathbb{1}(y_s^{(h,w,n)} == 1)
\label{eq:cfe}
\end{equation}
where $F(\cdot)$ is an CNN-Based feature extractor and $\mathbb{1}$ is the indicator function. Then, the naive representation of $n$ class prototype $p^n$ can be denoted as:
\begin{equation}
p^n = \frac{\sum_{x_s\in X_s}\sum_{h}\sum_{w} {f^n}^{(h,w)}}{\sum_{x_s \in X_s}\sum_h \sum_w \mathbb{1}(y_s^{(h,w,n)}==1)}
\label{Eq:pi}
\end{equation}
\par Meanwhile, this initial prototype is generally updated by Exponential Moving Average (EMA) model. Then, the prototypical alignment loss $\mathcal{L}_{a}$ is used as a auxiliary objective to pull the features of the $n$ class close to the related prototype $p^n$ in different ways, which can be generalized as:
\begin{equation}
\mathcal{L}_{a} = D(f^n, p^n)
\end{equation}
where $D(\cdot)$ is the distance function between the feature and related prototype, such as L2 distance and Kullback-Leibler divergence. Some works \cite{lai2022decouplenet,kundusubsidiary} pointed out that the shallow feature should be performed constraint for better distribution alignment. In other words, prototypical alignment is employed in multiple-layer features, called multi-layer prototypical alignment. The multi-layer prototypical alignment loss $\mathcal{L}_{ma}$ can be rewritten as:
\begin{equation}
\mathcal{L}_{ma} = \sum^L_l D(f_l^n, p^n)
\end{equation}
where $f_l$ is the feature extracted by the layer $l \in L$. For a clear comparison, Fig. \ref{Fig:PROTOCOMP} shows the architectures of the above methods. From the above description, there are two significant factors in prototypical alignment, i.e., prototype generation and alignment form. 
\par Nevertheless, few works discuss the rationality of the prototype generation and alignment form, especially in the DG task. It motivates us to investigate more effective ways for generalizable semantic segmentation from these two aspects.

\begin{figure*}[t]
	\centering
	\includegraphics[width=1\textwidth]{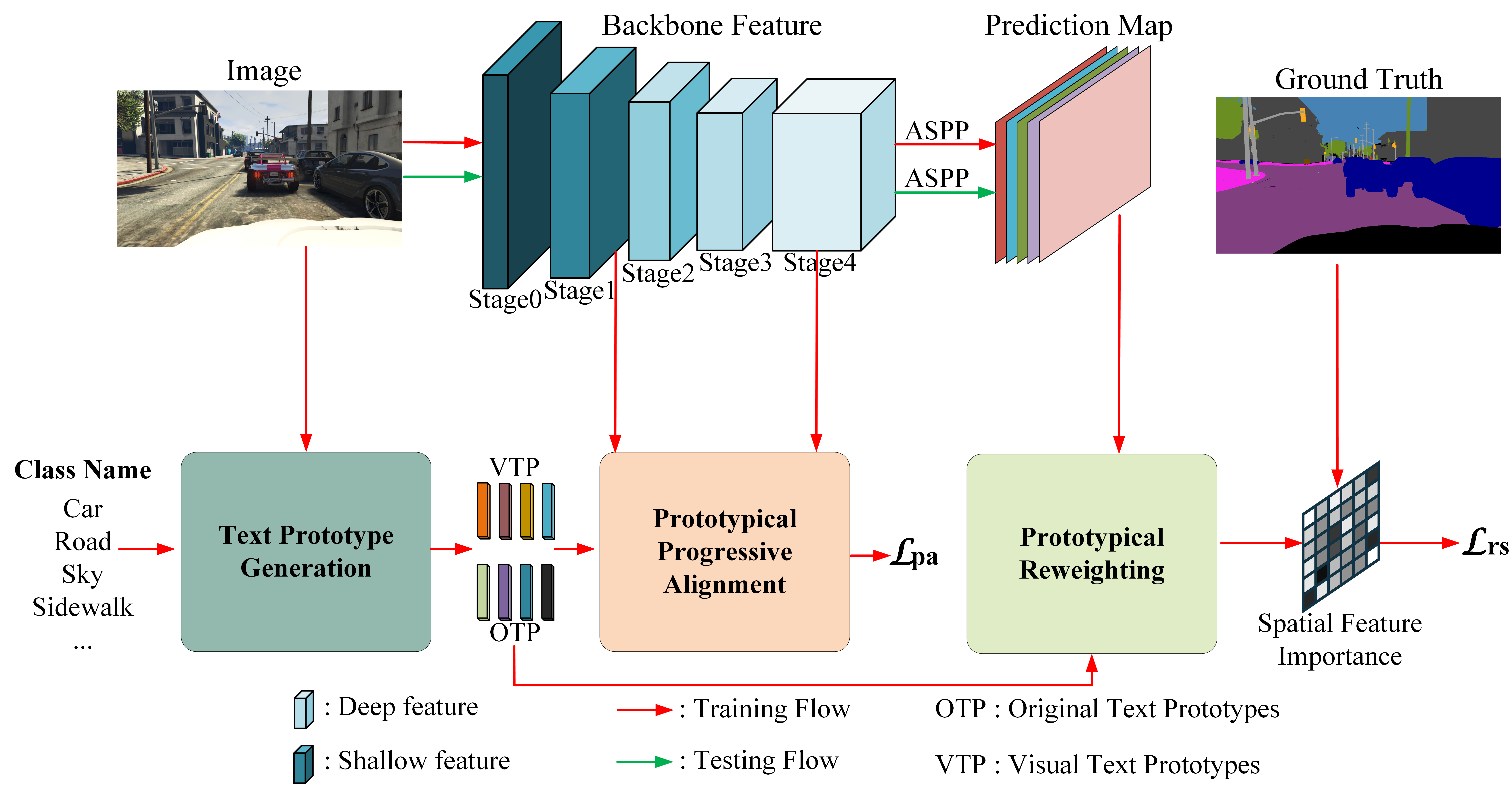}
	\caption{Flowchart of the proposed PPAR method including text prototype generation, prototypical progressive alignment, and prototypical reweighting. Blue cubes are features extracted from different layers. ASPP denotes Atrous Spatial Pyramid Pooling module \cite{chen2017rethinking}. The proposed modules are only used in the training stage to facilitate domain invariant representation generation.}
	\label{overview}
\end{figure*}
\section{Method}
\subsection{Framework overview}
The PPAR framework overview is shown in Fig. \ref{overview}, which consists of the text prototype generation, prototypical progressive alignment, and prototypical reweighting. These three modules are respectively and tightly related to the above three key observations in the Introduction Section. The overall objective $\mathcal{L}_{all}$ can be defined as:
\begin{equation}
\mathcal{L}_{all} = \mathcal{L}_{rs} + \alpha_{pa}\mathcal{L}_{pa}
\end{equation}
where $\mathcal{L}_{rs}$ and $\mathcal{L}_{pa}$ are the prototypical reweighting segmentation loss and prototypical progressive alignment loss, respectively. $\alpha_{pa}$ is the hyper-parameter as the weight of the progressive alignment loss during the training process.
% final objective.

\subsection{Text prototype generation}
In fact, there is a discrepancy between the class centroid of different domains \cite{lee2022bi}. Since the naive prototype calculated using Equation (\ref{Eq:pi}) is generated using the source domain data, which may lead to overfitting to the source domain \cite{cao2023adaptive}. As a result, such prototypical alignment pulls the feature close to the source prototype rather than the target prototype and cannot generalize well in possible target domains. Meanwhile, it may suffer from negative transfer \cite{wang2019characterizing} due to the influence of the outlier in the source data. There is no doubt that high-generalized and high-quality prototypes are beneficial for domain generalization. Inspired by other domain generalization methods \cite{yue2019domain,huang2021fsdr} that obtain a more generalized representation with the help of the wild (i.e., training on many large-scale auxillary datasets), our method adopts the CLIP model to generate text prototypes, as the CLIP model is trained on large-scale wild data and has strong generalizability \cite{radford2021learning}. The OTP $P^n_{to}$ can be calculated using the pred-trained text feature extractor $F_t(\cdot)$ in the CLIP model, which can be defined as:
\begin{equation}
P^n_{to} = F_t(T^n_o)
\label{eq:otpg}
\end{equation}
where $T_o^n$ is the string name of the $n$ class, such as ``car", ``person" and ``building".  After that, to perform prototypical alignment in a progressive manner, the VTP is proposed, which contains not only category information but basic visual information. Specifically, the basic visual information is injected into the OTP to generate the VTP.

\par In our case, color and texture are adopted as the basic visual factors. The color information $C^n$ of $n$ class is extracted as the mode (i.e., the color with the most frequency), which can be denoted as:
\begin{equation}
C^n =  Mo(I(x_s\mathbb{1}(y_s^{n} == 1)))
\end{equation}
where $Mo(\cdot)$ is the mode function and $I(\cdot)$ denotes the color value of the related pixel. Then, the texture information is extracted using the Local Binary Pattern $LBP(\cdot)$ \cite{ojala1994performance} since that effectively measures and extracts the local texture information of different classes. The Local Binary Pattern is widely used in computer vision tasks to improve classification performance \cite{pietikainen2011computer}. Similarly, the mode function is used to find the local texture information $LT^n$ with the most frequency, which can be denoted as:
\begin{equation}
LT^n =  Mo(LBP(x_s\mathbb{1}(y_s^{n} == 1)))
\end{equation}
\par After that, such basic visual factors are injected into the related class to generate text prototypes with basic visual information, i.e., VTP. Specifically, basic visual information represented as a string is concatenated with the text of the class name to generate visual text $T_v^n$, which is defined as:
\begin{equation}
T_v^n = \phi(T^n_o,``with\ color\ ", C^n, `` with\ local\ texture\ ", LT^n) 
\end{equation}
where $\phi$ represents the concatenation operation. Finally, the VTP $P^n_{tv}$ is also extracted by the pre-trained text feature extractor $F_t(\cdot)$, which can be denoted as:
\begin{equation}
P^n_{tv} = F_t(T^n_v)
\end{equation}
\par
\begin{figure}[t]
	\centering
	\includegraphics[width=0.48\textwidth]{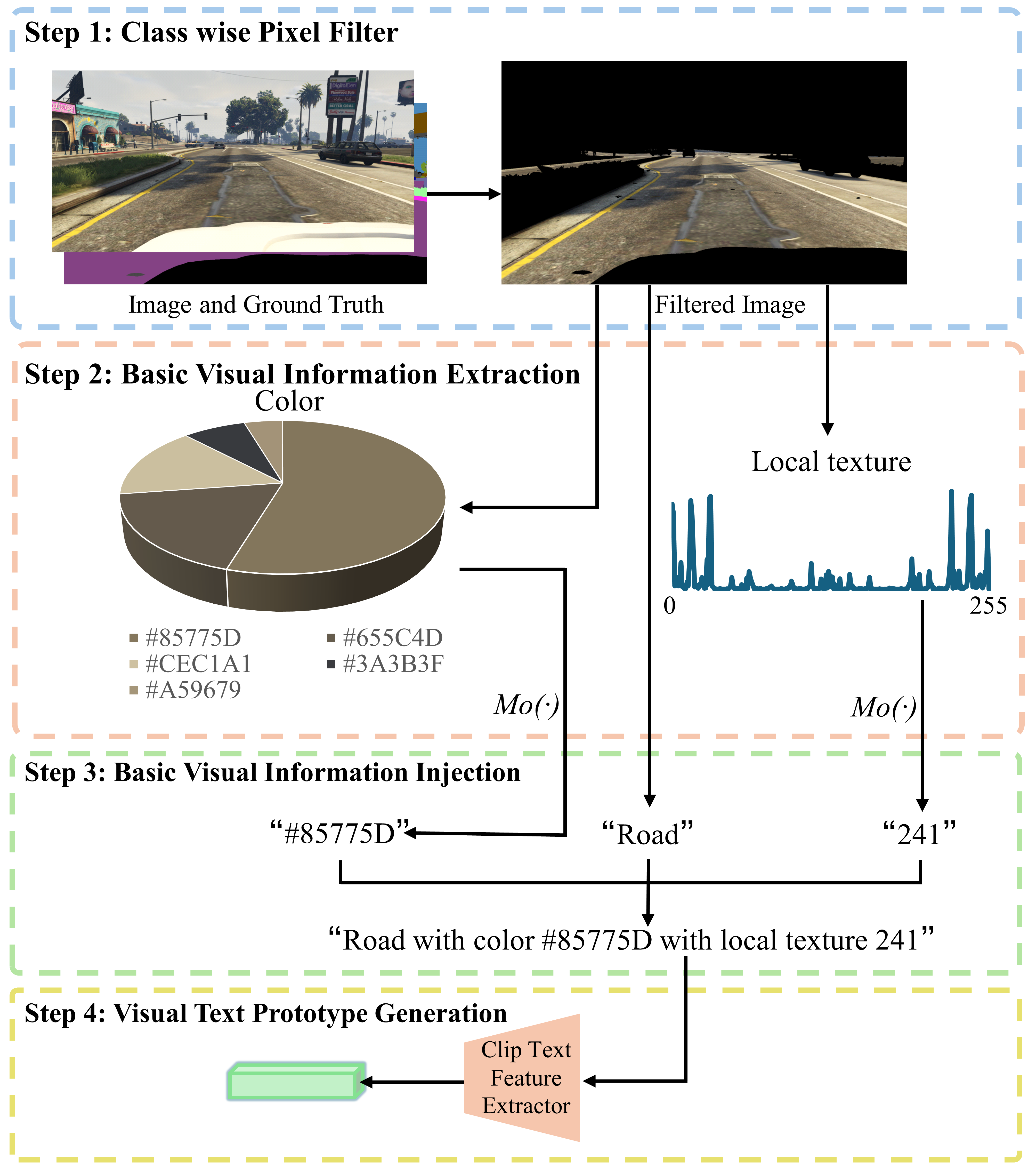}
	\caption{Illustration of VTP generation. The color and local texture information is first extracted and then injected into the OTP. In this example, color and local texture are represented as strings ``\#85775D" and ``241", respectively.}
	\label{Fig:BVI}
\end{figure}

 The basic visual information can be regarded as representative style information since the mode function is utilized to extract the most common color and local texture information of each category. The process of the visual texture prototype generation is illustrated in Fig. \ref{Fig:BVI} to understand clearly.

\subsection{Prototypical progressive alignment}
\begin{figure}[t]
	\centering
	\includegraphics[width=0.48\textwidth]{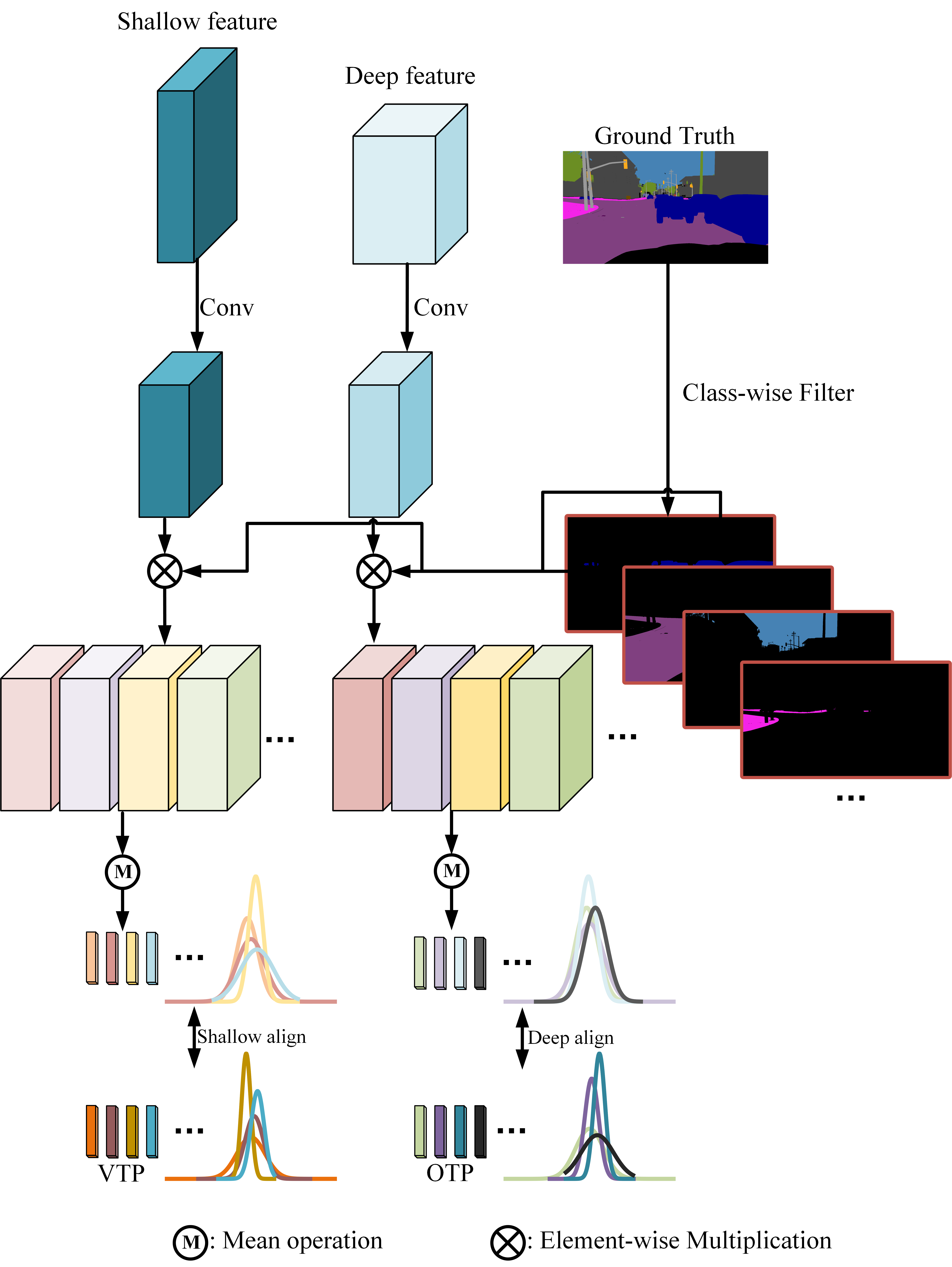}
	\caption{Illustration of prototypical progressive alignment. Class-wise features are first isolated based on the ground truth and text prototypes (OTP and VTP) are generated. Then, class-wise progressive alignment is performed, which utilizes text prototypes (OTP and VTP) to execute distribution alignment for better knowledge transfer.}
	\label{Fig:PPA}
\end{figure}

The next concern turned into the use of the class-wise text prototype.  According to the drawback of the existing methods in alignment form, the most important point is to design a refined multi-layer prototypical alignment form while considering the attributes of the shallow and deep features. As pointed out in some prior works \cite{gatys2016image, zhang2018fully}, the shallow feature is expected to carry more style information (contains basic visual information) while the deep feature is expected to carry more category information. Since the representative style information (i.e., basic visual information) is embedded into the OTP to generate VTP, a natural idea is to perform alignment utilizing the VTP for shallow features. Thus, the shallow feature can implicitly preserve the representative style information by shallow feature alignment. On the contrary, the deep feature is expected to contain category information for better classification, which matches that the OTP only contains generalized category information. It is reasonable to perform prototypical alignment using OTP for the deep features. Based on these analysis, the progressive alignment strategy is proposed, which is depicted in Fig. \ref{Fig:PPA}. Specifically, the proposed progressive alignment is performed at two features extracted in different layers, i.e., the shallow feature and the deep feature. The prototypical progressive alignment loss $\mathcal{L}_{pa}$ is divided into two parts:
\begin{equation}
\mathcal{L}_{pa} = \mathcal{L}_{as} + \mathcal{L}_{ad}
\end{equation}
where $\mathcal{L}_{as}$ and $\mathcal{L}_{ad}$ are shallow and deep prototypical alignment losses, respectively. For the shallow feature, a Kullback–Leibler divergence (KL) is adopted as the loss function to align the distribution between the VTP $P_{tv}$ and shallow class centroid $P_s$ calculated by Equation (\ref{Eq:pi}), which can be defined as:
\begin{equation}
\mathcal{L}_{as} = \sum_{n=0}^N  P^n_s \log \frac{P^n_s}{P^n_{tv}}
\end{equation}
\par For the deep feature, another KL loss between the OTP $P_{to}$ and deep class centroid calculated $P_d$ by Equation (\ref{Eq:pi}) is employed, which can be denoted as:
\begin{equation}
\mathcal{L}_{ad} = \sum_{n=0}^N P^n_d \log \frac{P^n_d}{P^n_{to}}
\end{equation}
\par To comprehend thoroughly this module, we reveal the insight behind the proposed method. The progressive alignment employs an easy-to-difficult alignment form, where the shallow feature alignment and the deep feature alignment can be treated as easy alignment and difficult alignment, respectively. For shallow feature alignment, given that VTP encapsulates both category and fundamental visual information about the source domain, shallow alignment can be interpreted as distribution alignment across the source domain and the intermediate domain between the source and pseudo overall domains. For the deep feature alignment, as OTP only contains category information, the alignment can be explained as the alignment between the source domain and the pseudo overall domains. Some evidence \cite{xu2020adversarial, wu2020dual, zhang2023hybrid} indicates that the introduction of the intermediate domain is more effective, thereby demonstrating that the former is easier to achieve the goal than the latter. Meanwhile, as features are progressively extracted from shallow to deep layers, our approach essentially follows a curriculum learning paradigm, learning knowledge from easy to difficult. Fortunately, the efficacy of such a curriculum learning approach has been validated by several UDA methods \cite{zhang2017curriculum, sakaridis2020map}. In summary, our method is elaborately designed by fully considering the attributes of the different features, meanwhile inspired by the curriculum learning and intermediate domain alignment, providing insight into the effectiveness of our approach.

\subsection{Prototypical reweighting}

\begin{figure}[t]
	\centering
	\includegraphics[width=0.48\textwidth]{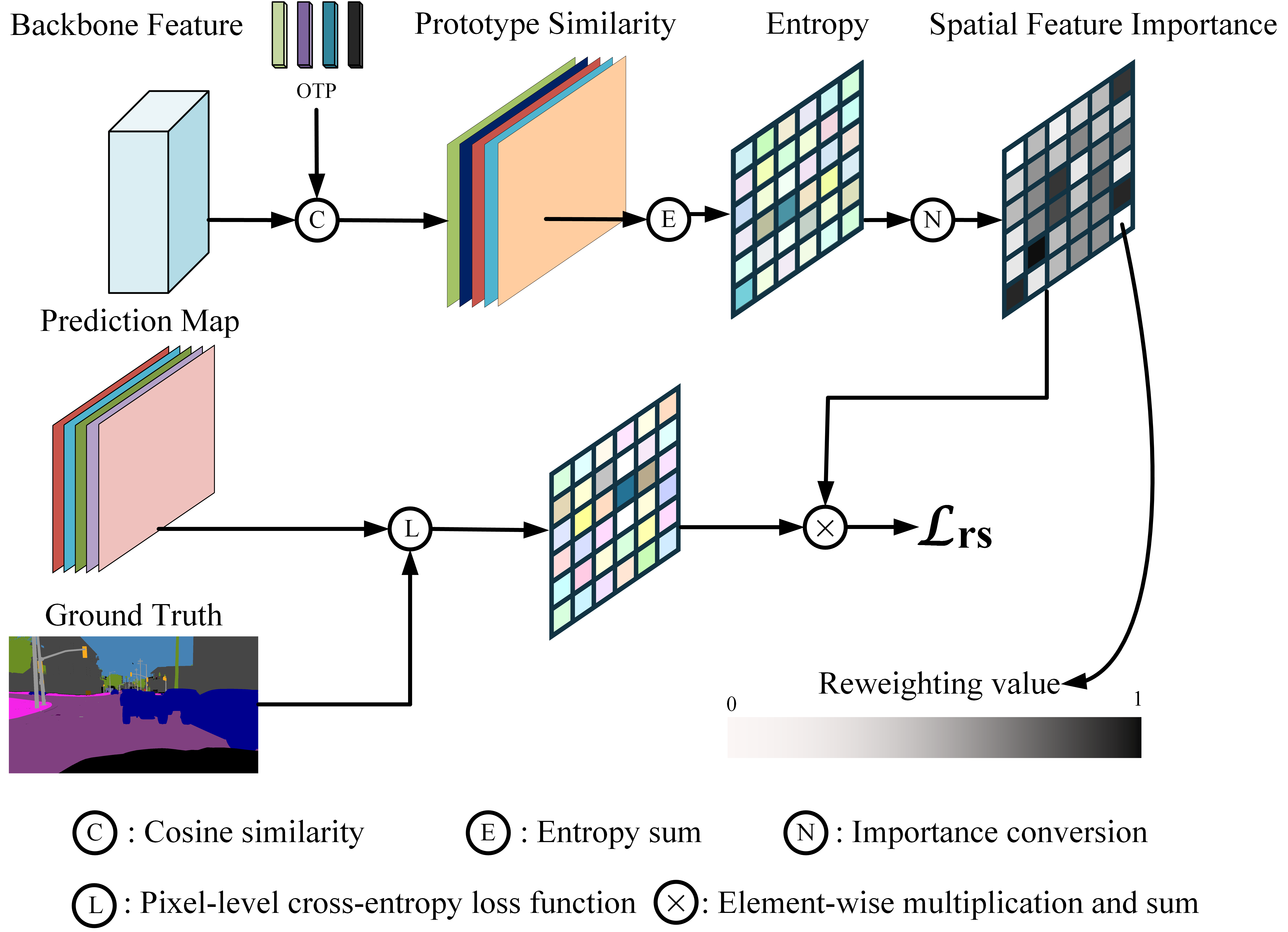}
	\caption{Illustration of prototypical reweighting. OTP is utilized as class-wise target centroids to estimate the feature importance. The feature importance is used as the reweighting information and embedded into the cross-entropy function to avoid learning outliers.}
	\label{Fig:PR}
\end{figure}

Besides avoiding the participation of irrelevant source features during prototype generation to deal with the overfitting problem and negative transfer, it is also necessary to consider this factor during feature learning \cite{zhang2024global}. Hence, we propose a prototypical reweighting learning strategy, which aims to estimate the importance of the source data and correct the learning weight of each source sample, as shown in Fig. \ref{Fig:PR}. Same as the prototypical progressive alignment, the proposed prototypical reweighting learning also relies on the text prototype due to its strong generalizability. In our proposed prototypical reweighting learning strategy, since semantic segmentation is a pixel-level task, the source importance should be furtherly reflected at the pixel level rather than at the image level. Considering the strong generalizability of the text prototype provided by the CLIP model, the similarity between feature and OTP is adopted to estimate feature importance, which can be presented as:
\begin{equation}
{S^n}^{(h,w)} = \frac{\exp(P^n_{to} \cdot f_l^{(h,w)})}{\sum^N_{i=1} \exp(P^i_{to} \cdot f_l^{(h,w)})}
\end{equation}
where ${S^n}^{(h,w)}$ is the similarity between the feature and prototype of $n$ class at the $(h,w)$ spatial location. $f_l$ is the feature extracted from the last layer of the feature extractor. Then, the prediction uncertainty $U^{(h,w)}$ is defined as the entropy of similarity for all classes, which can be denoted as:
\begin{equation}
U^{(h,w)} = -\sum_{n}^{N}{S^n}^{(h,w)}\log {S^n}^{(h,w)}
\end{equation}
A high uncertainty value represents a low correlation between the feature and the OTP. Next, the prediction uncertainty is normalized by linear normalization and is furtherly converted into the reweighting value $R^{(h,w)}$, which is denoted as:
\begin{equation}
R^{(h,w)} = \exp(-\frac{U^{(h,w)}-U_{min}}{U_{max}-U_{min}})
\end{equation}
where $U_{min}$ and $U_{max}$ respectively denote the minimum and maximum values of the prediction uncertainty $U$. Finally, the prototypical reweighting segmentation loss $\mathcal{L}_{rs}$ is defined as a reweigthing cross-entropy loss:
\begin{equation}\label{segloss} \mathcal{L}_{rs}=-\sum_{h,w} R^{(h,w)}\cdot \sum_{n\in N}y^{(h,w,n)}log(M^{(h,w,n)}) \end{equation}
where $M$ is the prediction map. The proposed strategy views the text prototype provided by the CLIP model as the class centroid of the unseen target domains since the CLIP model is trained with many large-scale auxiliary datasets that may cover most unseen target domains. Therefore, the source data with a low correlation to the OTP refers to unrelated data or outliers, which would be given a low learning weight to avoid forcibly learning from them. On the contrary, a feature with high similarity with the related class prototype is considered a more generalized feature and given a high weight to learn more generalized knowledge.

\subsection{Theoretical insight}
We also provide theoretical insight for our proposed method using domain adaptation/generalization theory \cite{ben2010theory, albuquerque2019generalizing}. Given a source domain $D_s \sim P_s$, a set of unseen target distributions $D_{ut} \sim P_{ut}$, a hypothesis $h \in \mathcal{H}$, since the source domain data is augmented by some photometric transformations to alleviate source overfitting problem like other DG methods \cite{choi2021robustnet,peng2022semantic}, the single source domain is extended to $N_s$ source domains $D_s^i \sim P_s^i, i \in N_s$. As pointed out in G2DM \cite{albuquerque2019generalizing}, the target distribution is a linear combination of all source distributions (i.e., $P_{ut}=\sum_{i=1}^{N_s}\alpha_i P_s^i, \sum_{i=1}^{N_s}\alpha_i = 1$). Furthermore, $\hat{P_{ut}}$ is introduced to describe the element contained in the convex hull of source domains which is closest to $P_{ut}$. The target risk $r_t(h)$ can be bounded by several terms:  
\begin{equation}
\begin{aligned}
r_t(h) &\leq \sum_{i=1}^{N_s}\alpha_i r_s^i(h) + \delta + \gamma \\ 
&+ \min{(\mathbb{E}_{\hat{P_{ut}}}[|f_{s_{\alpha}} - f_{ut}|],\mathbb{E}_{{P_{ut}}}[|f_{s_{\alpha}} - f_{ut}|])}
\end{aligned}
\end{equation}
where $r_s(h)$, $\delta$ and $\gamma$ are the source risk, the domain divergence between different source domains,  the domain divergence between $P_{ut}$ and $\hat{P_{ut}}$, respectively. $f_{s_{\alpha}}(x)=\sum_{i=1}^{N_s}\alpha_i f_{s}^i(x)$ is the label mapping function for any $x \in Supp(D_{ut})$ consisting of all $f_{s}^i$ with $\alpha_i$. Since the covariance shift assumption holds \cite{ben2010theory},  $f_{s_\alpha}$ equals $f_{ut}$ and then the rightmost term equals 0. Meanwhile, as the linear combination of source domains can represent the unseen target domain, the domain divergence between $P_{ut}$ and $\hat{P_{ut}}$ reduces to 0, i.e., $\gamma$ equal to 0. 

Consequently, the target risk $r_{ut}(h)$ is bounded by the source risk $r_s^i, i \in N_s$ and the domain divergence between diverse source domains $\delta$.  It means lower target risk with high generalizability can be achieved using lower $r_s^i, i \in N_s$ and $\delta$. First, for the $r_s^i, i \in N_s$, although previous supervised segmentation loss without the reweighting strategy can effectively reduce this risk, it suffers from the effect of out-of-distribution (OOD) samples that are generated by some photometric transformations \cite{lyu2022aadg}. That is, the source outlier or unrelated source data mentioned in the Introduction Section. In our case, the proposed prototypical reweighting supervised loss embedded the feature weights into a cross-entropy function in Equation (\ref{segloss}), which decreases the learning weight value for these outliers to alleviate the effect of these OOD data and obtain a lower source risk. Second, for the $\delta$, as discussed above, different source domains can be seen as the effect of data augmentation operation. Thus, there is a distribution gap between inter-source domains. The proposed progressive alignment strategy adopts the text prototype that is close to the pseudo overall domain, and performs alignment between the distribution of the pseudo overall domain and that of diverse source domains, achieving the goal of minimizing $\delta$. In summary, the proposed approach complies well with domain generalization theory \cite{albuquerque2019generalizing}.
\section{Experiment}
In this section, extensive experiments of our approach are conducted on several challenging datasets. Further, three backbone architectures are adopted and the segmentation visualization is provided.

\begin{table*}[t]
	\caption{Performance comparison with other SOTA methods using ResNet-50 backbone. \textbf{Bold} format is the best performance and \uline{Underline} is the second-best performance. Avg represents the average mIoU of all evaluation settings. Methods marked $^\dagger$ report reproductive performance due to some performance lack in their related papers. ``-" indicates the absence of results.}
	
	\label{tab:DG performance}
	\centering{}\resizebox{1.0\textwidth}{!}{%
		\begin{tabular}{ccc|cccc|cccc|cccc|cccc|cccc}
			\toprule 
			\multirow{2}{*}{Model} &\multirow{2}{*}{Methods}&   \multirow{2}{*}{Avg}& \multicolumn{4}{c}{Trained on GTA5 (G)} &  \multicolumn{4}{c}{Trained on SYNTHIA (S)} &   \multicolumn{4}{c}{Trained on Cityscapes (C)} &  \multicolumn{4}{c}{Trained on BDD100K (B)} &  \multicolumn{4}{c}{Trained on Mapillary (M)} \\
			&& & $\rightarrow$C & $\rightarrow$B & $\rightarrow$M & $\rightarrow$S & $\rightarrow$C & $\rightarrow$B & $\rightarrow$M & $\rightarrow$G &  $\rightarrow$B & $\rightarrow$M & $\rightarrow$G & $\rightarrow$S &  $\rightarrow$G & $\rightarrow$S & $\rightarrow$C & $\rightarrow$M &  $\rightarrow$G & $\rightarrow$S & $\rightarrow$C & $\rightarrow$B  \tabularnewline 
			\hline 
			\noalign{\vskip0.1cm}
			&IBN \cite{pan2018two} &  34.2 & 33.9 & 32.3 & 37.8 & 27.9 &  32.0 & 30.6 & 32.2 & 26.9  & 48.6 & 57.0 & 45.1 & 26.1 & 29.0 & 25.4& 41.1 & 26.6 &  30.7 & 27.0 & 42.8 & 31.0\tabularnewline
			&SW \cite{pan2019switchable} &   32.2 & 29.9 & 27.5 & 29.7 & 27.6  & 28.2 & 27.1 & 26.3 & 26.5 &  48.5 & 55.8 & 44.9 & 26.1&  27.7 & 25.4 & 40.9 & 25.8 & 28.5 & 27.4 & 40.7 & 30.5\tabularnewline
			&DRPC \cite{yue2019domain} &   35.8 & 37.4 & 32.1 & 34.1 & 28.1 &35.7 & 31.5 & 32.7 & 28.8 &  49.9 & 56.3 & 45.6 & 26.6 & 33.2 & 29.8 & 41.3 & 31.9 &  33.0 & 29.6 & 46.2 & 32.9\tabularnewline
			&GTR \cite{peng2021global}&   36.1 & 37.5 & 33.8 & 34.5 & 28.2 & 36.8 & {32.0} & {32.9} & 28.0 & {50.8} & 57.2 & {45.8} & 26.5 & 33.3 & 30.6 & 42.6 & 30.7  & 32.9 & 30.3 & 45.8 & 32.6 \tabularnewline
			&ISW \cite{choi2021robustnet} &   36.4 & 36.6 & 35.2 & 40.3 & 28.3 &  35.8 & 31.6 & 30.8 & 27.7  & 50.7 & 58.6 & 45.0 & 26.2 &  32.7 & {30.5} & 43.5 & 31.6 & 33.4 & 30.2 & 46.4 & 32.6\tabularnewline
			&SAN-SAW \cite{peng2022semantic} &   {38.5} &  {39.8} & {37.3} & {41.9} & {30.8}  & {38.9} & \textbf{35.2} & {34.5} & {29.2} &   \textbf{53.0} & \textbf{59.8} & {47.3} & {28.3}  & {34.8} & \uline{31.8} & {44.9} &{33.2}&  {34.0} & {31.6} & {48.7} & {34.6} \tabularnewline[0.1cm]
			&SIL \cite{zhang2023learning} &   {40.4} & {41.5} & {37.8} &{43.4}&\underline{33.5}&{39.2}&30.4&31.0&{34.8}&{50.8}&\uline{59.2}&44.8&{30.9}&{40.1}&26.2&{50.9}& {49.1} & {39.6}  & {30.9}& {50.4}& {44.1}   \tabularnewline[0.1cm]
			&PinMem $^\dagger$ \cite{kim2022pin} &   41.0 &41.2 &35.2 &39.4&28.9& 38.2& {32.3} &33.9 &32.1& 50.6 &57.9& 45.1& 29.4& 42.4 &29.1 &54.8 &51.0 &44.1& 30.8 &55.9& 47.6 \tabularnewline[0.1cm] 
			%DIRL \cite{xu2022dirl}&   41.1 & \textbf{41.0} & \textbf{39.2} & 41.6 & - & - & - & - & - & {51.8} & - &  {46.5} & 26.5 & -&- &- & -& -&- & - & - \tabularnewline[0.1cm]
			%WildNet\cite{lee2022wildnet}&   43.1 &  44.6 & 38.4 &  46.1 & 31.3 & - & - & - & - &  50.9 & 58.8 & 47.0 & 28.0 & -&- &- & -& -&- & - & - \tabularnewline[0.1cm]
			ResNet-50&NSFRC \cite{zhang2024video}  &   {42.2} & {42.6} & {37.9} &{42.0}&{33.1}&\uline{39.5}&30.0&32.3&{29.7}&{50.9}&57.8&45.3&{30.5}&30.6&\textbf{33.6}&\textbf{57.4}& \textbf{56.2} & \textbf{49.7}  &\uline{34.6}& \textbf{60.0}& \textbf{51.2}   \tabularnewline[0.1cm]
			&DIIA \cite{liao2023domain} &   42.3  & 41.6 &38.6& 42.8& 32.1 &38.4& \uline{35.0}& \textbf{35.8}& \uline{35.5}& \uline{52.7} &58.8 &\uline{47.7} &\textbf{32.3} &44.1 &30.6 &52.0 &50.8 &44.2 &31.9 &54.3 &46.5 \tabularnewline[0.1cm]
			&SHADE $^\dagger$  \cite{zhao2022style} &   42.5 & \underline{43.5} &\textbf{40.3} &{43.0} & {31.2} & \textbf{39.6} &29.2 & \uline{34.7} &{34.8} &{51.5} & {58.7} & \textbf{48.2} & {30.8} & {43.5} & 31.1 & \uline{56.2} & {53.1} & {45.3} & 31.1 & {56.2} & {48.5} \tabularnewline[0.1cm]
			&FGSS \cite{zhang2023fine}  &   \underline{42.9} & \textbf{44.5} & {36.5} &\uline{44.3}&{31.4}&{39.2}&29.8&33.5&{34.5}&{52.0}&58.1&45.1&{30.0}&\uline{47.0}&{30.8}&\uline{56.2}& \uline{55.5} & {46.7}  & \textbf{35.5}& \uline{57.4}& {49.8}  \tabularnewline[0.1cm]
			
			%&VRT \cite{yue2023visual} &   - & 34.6& 34.5 &39.3 &30.0 &-&-&-&-&47.2 & 56.5 & 44.6& 27.7&-&-&-&-&-&-&-&-\tabularnewline[0.1cm]
			&NP \cite{fan2023towards} & -&40.6& 35.6& 38.9& 27.7 & - & - &- &- &- &- &- &-&-&-&-&-&- &-&-&- \tabularnewline[0.1cm]
			&ProRandConv \cite{choi2023progressive} &   - & 42.4& 38.8& 41.6 &25.5 & - &- &- &- &- &- &-&-&-&-&-&-&-&-&-&- \tabularnewline[0.1cm]
			&ITSA \cite{chuah2023information} &   - & 41.0& 36.5 &42.3 &-& -& - &- &- &- &- &- &-&-&-&-&-&- &-&-&- \tabularnewline[0.1cm]
			&OCR \cite{jing2023order} &   - &38.9&37.2 & 39.7 & 27.0 &-& -& - &- &- &- &- &- &-&-&-&-&-&- &-&-\tabularnewline[0.1cm]
			
			&Ours &   \textbf{43.1} & {42.7} & \uline{39.7} &\textbf{44.5}&\textbf{33.6}&{38.9}&31.6&30.8&\textbf{37.5}&{51.1}&{58.4}&45.2&\uline{31.7}&\textbf{47.3}&{31.3}&{54.9}& {54.2} & \uline{47.4}  & {34.1}& {56.9}& \uline{49.9}  \tabularnewline[0.1cm]
			
			\hline
			\noalign{\vskip0.1cm}
	\end{tabular}}
	
\end{table*}

\subsection{Datasets} 
Six datasets including Cityscapes \cite{cordts2016cityscapes}, BDD100K \cite{yu2020bdd100k}, Mapillary  \cite{neuhold2017mapillary}, IDD \cite{varma2019idd}, SYNTHIA  \cite{ros2016synthia}, and GTAV \cite{richter2016playing} are adopted for training and evaluation, where the first four datasets are captured in the real world and others are synthetic. 
\par For real-world datasets, the Cityscapes dataset has 2975 training and 500 validation images with a resolution of 2048 $\times$ 1024. The BDD100K dataset contains 7000 training and 1000 validation images with a resolution of 1280 $\times$ 720. The Mapillary dataset is a large-scale street scene dataset containing 18000 training and 2000 validation images that have a resolution of at least 1920$\times$1080. The IDD dataset is captured in a complicated Indian environment, including 10004 images with a resolution of 1678 $\times$ 968.
\par For synthetic datasets, the GTAV dataset is captured by the game Grand Theft Auto V containing 24996 images with a resolution of 1914 $\times$ 1052. The SYNTHIA dataset is captured by the simulator from multiple views, which consists of 9400 images with a resolution of 1280 $\times$ 760. The abbreviations G, S, C, B, M, and I represent the GTAV, SYNTHIA, Cityscapes, BDD100K, Mapillary, and IDD, respectively.
\subsection{Implementation}
ResNet-50 \cite{he2016deep}, MobileNet-V2 \cite{sandler2018mobilenetv2}, and ShuffleNet-V2 \cite{ma2018shufflenet} are adopted as the feature extractor of the segmentation model. All backbones are initialized with the pre-trained model of the ImageNet. Then an Atrous Spatial Pyramid Pooling (ASPP) \cite{chen2017rethinking} module is utilized as the classifier of the segmentation model like most DG methods \cite{choi2021robustnet, peng2022semantic}. An SGD optimizer is used to optimize the segmentation model with an initial learning rate of 0.01, a momentum of 0.9, and a weight decay of 0.0025. The batch size is set to 4. The image resolution of 768$\times$768 is the cropped training resolution. Following the standard semantic segmentation task, the mean Intersection Over Union (mIoU) is adopted as the evaluation metric. The max iteration is 100000 but the training program early stops in 60000 iterations. Pytorch library is utilized to implement the proposed method. The Stage 1-4 features represent the \{conv2\_x, conv3\_x, conv4\_x, conv5\_x\} features in the ResNet architecture. $\alpha_{pa}$ is set to 0.001. 

\begin{table}[t]
	\caption{Performance comparison with other SOTA methods using MobileNet-V2 and ShuffleNet-V2 backbones on G $\rightarrow$ \{C, B, M\} and C $\rightarrow$ \{G, B, S\} generalized task.}
	\label{tab:DG_GC_to_Other} 
	
	\centering{}\doublerulesep=0.5pt \resizebox{0.5\textwidth}{!}{
		\begin{tabular}{ccc|ccc|ccc}
			%\toprule[0.2em]
			\toprule 
			\multirow{2}{*}{Model} & \multirow{2}{*}{Method} &  \multirow{2}{*}{Avg} & \multicolumn{3}{c}{Trained on G}  &   \multicolumn{3}{c} {Trained on C}  \\
			&&& $\rightarrow$C&  $\rightarrow$B &  $\rightarrow$M  &$\rightarrow$B& $\rightarrow$S &  $\rightarrow$G  \tabularnewline
			%\midrule[0.15em] 
			\cline{1-9}
			\multirow{9}{*}{ShuffleNet-V2} %& Baseline & 25.6 & 22.2 & 28.6 & 25.4 \tabularnewline
			& IBN\cite{pan2018two} &33.3& 27.1 & 31.8 & 34.9 & 41.9 & 23.0 & {40.9}  \tabularnewline
			& ISW \cite{choi2021robustnet}&33.9& 31.0 & 32.1 & 35.3 & 41.9 & 22.8 & 40.2  \tabularnewline
			& DIRL  \cite{xu2022dirl}&34.7& {31.9} & {32.6} & {36.1} & {42.6} & {23.7} & \uline{41.2} \tabularnewline
			& SAN-SAW \cite{peng2022semantic} &34.3&31.9 &30.2 &34.8 & 42.3 &\uline{27.6}& 38.8  \tabularnewline
			& SIL \cite{zhang2023learning} &35.9& {34.5} & \uline{33.4} & {36.6}& {43.9}& {26.1} & {40.6}  \tabularnewline
			& PinMem $^\dagger$ \cite{kim2022pin} &33.2& 29.5& 31.3& 35.4 &39.9 & 25.9 & 37.4  \tabularnewline
			& SHADE $^\dagger$  \cite{zhao2022style} &36.0& \uline{35.4} & 32.3 & {36.9} & \uline{44.3} & {26.4} & {40.9}  \tabularnewline 
			& FGSS \cite{zhang2023fine} &{36.3}& {35.3} & {33.1} & \uline{37.6} & {44.1}& {27.5} & {40.3}   \tabularnewline
			& DIIA \cite{liao2023domain}  & \uline{36.8} & 34.6 &\uline{33.4} &37.5 & \textbf{44.5} &\uline{27.6} & \textbf{43.1}  \tabularnewline
			& Ours &\textbf{37.4}& \textbf{35.8} & \textbf{35.1} & \textbf{39.3} & {42.8}& \textbf{31.9} & {39.6}   \tabularnewline
			
			\cline{1-9}
			%\midrule[0.15em] 
			\multirow{9}{*}{MobileNet-V2} % & Baseline & 25.9 & 25.7 & 26.5 & 26.0\tabularnewline
			& IBN \cite{pan2018two}  &32.4& 30.1 & 27.7 & 27.1 & 45.0 & 23.2& 41.1  \tabularnewline
			& ISW \cite{choi2021robustnet}&33.5& 30.9 & 30.1 & 30.7 & 45.2 & 22.9 & 41.2  \tabularnewline
			& DIRL  \cite{xu2022dirl}&35.7& {34.7} & {32.8} & {34.3} & \textbf{47.6} & 23.3 & {41.4}  \tabularnewline
			& SAN-SAW \cite{peng2022semantic} &34.1& 32.5& 27.6 & 30.8 & 45.8 & 26.7 &41.2 \tabularnewline
			& SIL \cite{zhang2023learning} &36.9& \uline{36.3} & {33.0} & \textbf{36.2}& {46.3} & {27.2} & \uline{42.2}  \tabularnewline
			& PinMem $^\dagger$ \cite{kim2022pin} &34.4& 32.2& 29.0& 31.5 & {46.2} & 26.8 & 40.4  \tabularnewline
			& SHADE $^\dagger$  \cite{zhao2022style} &35.6& 34.4 & 32.4 & 32.4 & 46.1 & {27.2} & {41.2}  \tabularnewline 
			& FGSS \cite{zhang2023fine} &36.9& {35.8} & {33.2} & \uline{36.0} & \uline{46.6} & \uline{27.6} & \textbf{42.3}  \tabularnewline
			& DIIA \cite{liao2023domain} & \uline{37.1} & \textbf{37.0}& \uline{33.6} &34.7& 47.8 & 27.6& 42.1 \tabularnewline
			& Ours &\textbf{38.2}& {36.2} & \textbf{34.5} & {35.7} & \textbf{47.6} & \textbf{33.5} & {41.4}  \tabularnewline
			\bottomrule
			%\bottomrule[0.15em]
	\end{tabular}}
\end{table}
\subsection{Comparison with state-of-the-art methods}
To comprehensively verify the generalizability of the DG model, several state-of-the-art (SOTA) methods are compared in three backbones.
\subsubsection{Performance comparison using ResNet-50 backbone} 
For the ResNet-50 backbone, we compared 16 representative methods. We adopted the same setting as SAN-SAW \cite{peng2022semantic} and there are five evaluation settings in our experiments, including G $\rightarrow$ \{B, M, S, C\}, S $\rightarrow$ \{G, C, B, M\}, C $\rightarrow$ \{G, S, B, M\}, B $\rightarrow$ \{G, S, C, M\}, and M $\rightarrow$ \{G, S, B, C\}, where $\rightarrow$ represents ``generalizing to". The average mIoU across all target domains is adopted as the final evaluation metric. As shown in Table \ref{tab:DG performance}, our proposed method achieves an average mIoU of 43.1\%,  surpassing several SOTA methods including SAN-SAW \cite{peng2022semantic}, SIL\cite{zhang2023learning}, PinMem\cite{kim2022pin}, and SHADE\cite{zhao2022style}, with respective improvements of 4.6\%, 2.7\%, 2.1\%, and 0.6\% in mIoU. Compared to our baseline ISW \cite{choi2021robustnet}, our PPAR method has a clear improvement of 6.7\% in terms of average mIoU. Although FGSS \cite{zhang2023fine} has a similar performance to our method with the ResNet-50 backbone, our method outperforms FGSS with a gain of 1.1\% and 1.3\% in mIoU with the ShuffleNet-V2 and MobileNet-V2 backbones, respectively, shown in Table \ref{tab:DG_GC_to_Other}. Besides the average mIoU, the frequency of the first/second best also serves as an indicator of model generalizability. Our method obtained first/second best mIoU in 8 instances, outperforming other methods. Although NSFRC \cite{zhang2024video} shares the same best number as our method, it reveals a serious domain imbalance problem in the B $\rightarrow$ \{G\} setting. For some methods that only provided the experimental result of G $\rightarrow$ \{C, B, M, S\}, i.e., NP \cite{fan2023towards}, ProRandConv \cite{choi2023progressive}, ITSA \cite{chuah2023information}, and OCR  \cite{jing2023order}, there is a clear margin between their methods and ours.
\subsubsection{Performance comparison using other backbones} 	
For MobileNet-V2 and ShuffleNet-V2, following the ISW \cite{choi2021robustnet} and DIRL \cite{xu2022dirl}, experiments on two evaluation settings containing G $\rightarrow$ \{C, B, M\} and C $\rightarrow$ \{G, S, B\} are conducted shown in Table \ref{tab:DG_GC_to_Other}. Compared to 9 SOTA methods,  the proposed method achieves the best performance of 37.4\% and 38.2\% in terms of average mIoU in the ShuffleNet-V2 and MobileNet-V2 backbone, respectively. With these performances, our method has a gain with at least 0.6\% and  1.1\% in mIoU compared to other methods, respectively. These results show the superiority of our PPAR method. 
\subsubsection{Performance comparison on multi-source domain generalization}
In addition, we conducted two multi-source domain generalization experiments. As shown in Table \ref{table:multisourcemix}, our method, training the model on GTAV and SYNTHIA datasets and testing it on Cityscapes, BDD100K, and Mapillary datasets, achieved a significant improvement over the baseline ISW, with a gain of 5.5\% in mIoU. Even when PinMem \cite{kim2022pin}  is a framework designed for multi-source domain generalization, our method still outperforms it by 0.5 \% in mIoU performance. In the task of {G + S + I $\rightarrow$ \{C, B, M\}}, as shown in Table \ref{table:multisourceGSImix}, our method also surpasses the PinMem \cite{kim2022pin} with a clear increase of 1.1 \% in average mIoU. These results show the effectiveness of our method in multi-source domain generalization.
\begin{table}[t]
	\centering
	\tabcolsep=0.2cm
	\caption{Performance comparison on multi-source domain generalization, i.e., G + S $\rightarrow$ \{C, B, M\}.}
	\small
	\begin{tabular}{l|cccc}
		\bottomrule
		Method & $\rightarrow$C & $\rightarrow$B & $\rightarrow$M  & Avg  \\
		\cline{1-5}
		DeepLab-V3 \cite{chen2018encoder}&35.5 &25.1 &31.9 &30.8  \\
		IBN \cite{pan2018two}& 35.6 &32.2 &38.1& 35.3 \\
		ISW\cite{choi2021robustnet} & 37.7& 34.1 &38.5& 36.8 \\ 
		MLDG \cite{li2018learning} &38.8& 32.0 &35.6 &35.5 \\
		TSN \cite{zhang2022generalizable} & 38.1 & - & - & - \\
		PinMem \cite{kim2022pin} &\textbf{44.5} &\uline{38.1}& \uline{42.7} &\uline{41.8} \\
		Ours & \uline{43.8} & \textbf{38.8} & \textbf{44.3} & \textbf{42.3}  \\
		\toprule
	\end{tabular}
	\label{table:multisourcemix}
\end{table}

\begin{table}[t]
	\centering
	\tabcolsep=0.2cm
	\caption{Performance comparison on multi-source domain generalization, i.e., G + S + I $\rightarrow$ \{C, B, M\}.}
	\small
	\begin{tabular}{l|cccc}
		\bottomrule
		Method & $\rightarrow$C & $\rightarrow$B & $\rightarrow$M  & Avg  \\
		\cline{1-5}
		DeepLab-V3 \cite{chen2018encoder}&52.5 &47.5 &54.7 &51.6  \\
		IBN \cite{pan2018two}& 54.4 &48.9 &56.1& 53.1\\
		ISW\cite{choi2021robustnet} & 54.7& 49.0 &56.9& 53.5 \\ 
		MLDG \cite{li2018learning} &54.8& 48.5 &55.9 &53.1 \\
		TSN \cite{zhang2022generalizable} & 53.0 & 46.4 & 52.8 & 50.7 \\
		PinMem \cite{kim2022pin} &\uline{56.6} &\uline{50.2}& \uline{58.3} &\uline{55.0} \\
		Ours &  \textbf{57.4} & \textbf{51.5} & \textbf{59.3} & \textbf{56.1}   \\
		\toprule
	\end{tabular}
	\label{table:multisourceGSImix}
\end{table}

\subsubsection{Performance with DGSS models using the Visual Foundation Model (VFM)}
Further comparison with recent VFM-based methods is crucial, as our method also involves the utilization of the foundation model (i.e., Text Foundation Model (TFM)). Both Rein \cite{wei2024stronger} and Tqdm \cite{pak2024textual} represent current SOTA approaches. Different from our proposed method that performs a knowledge transfer from text features generated by TFM to the visual feature extracted by small backbones, they employ VFM as backbones (Dinov2 \cite{oquab2023dinov2}, ViT \cite{radford2021learning}, Eva-02 \cite{sun2023eva}) and perform fine-tuning. Thus, we have included them as baselines to ensure a fair comparison and evaluated our approach under identical conditions. As shown in Table \ref{table:vfmcomp}, our approach (Rein + Ours) achieves an improvement over Rein, particularly in the G$\rightarrow$C and G$\rightarrow$M tasks. The average mIoU across all target domains for our method is 64.8\%, which is a 0.5\% improvement over Rein’s 64.3\%. More significantly, when compared to Tqdm with the ViT backbone, our method achieves an average mIoU of 56.3\%, outperforming Tqdm by 1.3\%. Specifically, we observe improvements of 0.5\%, 1.6\%, and 2.0\% in the G$\rightarrow$C, G$\rightarrow$B, and G$\rightarrow$M tasks, respectively. When compared to Tqdm using the Eva-02 backbone, our approach achieves an average mIoU of 67.1\%, outperforming Tqdm by 1.0\%. Specifically, we observe improvements of 0.4\%, 1.6\%, and 1.0\% in the G$\rightarrow$C, G$\rightarrow$B, and G$\rightarrow$M tasks, respectively. Meanwhile, it can be seen that our method leads KTBOS\cite{luo2025kill} (2025 IJCV) and SBS\cite{hummer2024strong} (2024 ACCV) by 6.4\% and 0.3\% in terms of average mIoU using ViT backbone. Meanwhile, our method outperforms SBS \cite{hummer2024strong} with a gain of 3.9\% in average mIoU using Eva-02 backbone.
\par Compared to our previous works (CSDS \cite{liao2024class} and CMPCL \cite{liao2024calibration}) published in TITS in 2024, our method still shows comparative performance as demonstrated in Table \ref{table:vfmcomp}. Specifically, compared to CSDS, our method outperforms DAformer-based CSDS by 1.1\% in average mIoU. In G$\rightarrow$C and G$\rightarrow$M settings, it achieves gains of 0.8\% and 3.4\%, respectively. Furthermore, compared to CMPCL \cite{liao2024calibration}, our method using Dinov2 backbone achieves an average mIoU of 64.8\%. Additionally, it shows improvements of 2.3\%, 0.5\%, and 0.2\% in G$\rightarrow$\{C, B, M\}, compared to CMPCL equipped with the Dinov2 backbone. 
\par These experimental results show that our method helps domain-invariant feature extraction not only of small backbones (e.g., ResNet-50, MobileNet, ShuffleNet) still of current VFM models (Dinov2 \cite{oquab2023dinov2}, ViT \cite{radford2021learning}, and Eva-02 \cite{sun2023eva}). It presents great scalability and generalization of our approach.
\begin{table*}[t]
	\centering

	\caption{Performance comparison with DGSS models using the VFM.}
	\small
	\begin{tabular}{c|c|c|cccc}
		\bottomrule
		Backbone & Method  &Venue &  $\rightarrow$C & $\rightarrow$B & $\rightarrow$M  & Avg  \\
		\cline{1-7}
		& DAFormer \cite{hoyer2022daformer} & CVPR2022 &52.7 & 47.9 & 54.7 & 51.8 \\ 
		& HRDA \cite{hoyer2023domain}& TPAMI2023 & 57.4 & 49.1 & 61.2 & 55.9 \\

		&	KTBOS \cite{luo2025kill} & IJCV2025 & 49.6 & 48.4 & 51.8 & 49.9 \\
		& SHADE \cite{zhao2024style} & IJCV2024 & 53.3  &48.2 &55.0 &52.2\\
		&	Tqdm \cite{pak2024textual} & ECCV2024 & 57.5 & 47.7 &59.8 &55.0 \\
		ViT & DAFormer + CSDS \cite{liao2024class} & TITS2024& 57.2 & {49.5} & 58.4 & 55.0 \\
		& CMFormer \cite{bi2024learningaaai} & AAAI2024 & 55.3& 49.9& 60.1& 55.1 \\
		& SBS \cite{hummer2024strong} & ACCV2024 & 55.6 &\textbf{52.5} &59.9 & 56.0\\ 
		& HRDA + CSDS \cite{liao2024class}& TITS2024 & \textbf{60.9} & 50.2 & \uline{63.1} & \uline{58.1} \\
		&  CMPCL \cite{liao2024calibration} & TITS2024 & \uline{59.4} & \uline{52.2} & \textbf{63.4} & \textbf{58.3} \\
		& Ours  & & {58.0} & {49.3} &{61.8} & {56.3} \\ 
		\cline{1-7}
		& CMPCL \cite{liao2024calibration}  & TITS2024& 65.5 & 59.8 & 66.0 & 63.8 \\
		Dinov2	&	Rein \cite{wei2024stronger} & CVPR2024 & 66.4 &\textbf{60.4} &\uline{66.1}& \uline{64.3}   \\

		&  Ours &  & \textbf{67.8}& \uline{60.3}& \textbf{66.2} & \textbf{64.8}  \\
		
		\cline{1-7}

		& SBS\cite{hummer2024strong} & ACCV2024 & 65.3 &58.3& 66.0 & 63.2 \\
		Eva-02	&	Rein \cite{wei2024stronger} & CVPR2024 & 65.3 & \uline{60.5} & 64.9& 63.6 \\
		& 	Tqdm  \cite{pak2024textual} & ECCV2024&  \uline{68.9} & 59.2 &\uline{70.1} & \uline{66.1} \\
		& Ours & & \textbf{69.3} & \textbf{60.8} &\textbf{71.1} & \textbf{67.1} \\ 
		\toprule
	\end{tabular}
	\label{table:vfmcomp}
\end{table*}

\begin{figure*}[t]
	\centering
	\includegraphics[width=0.98\textwidth]{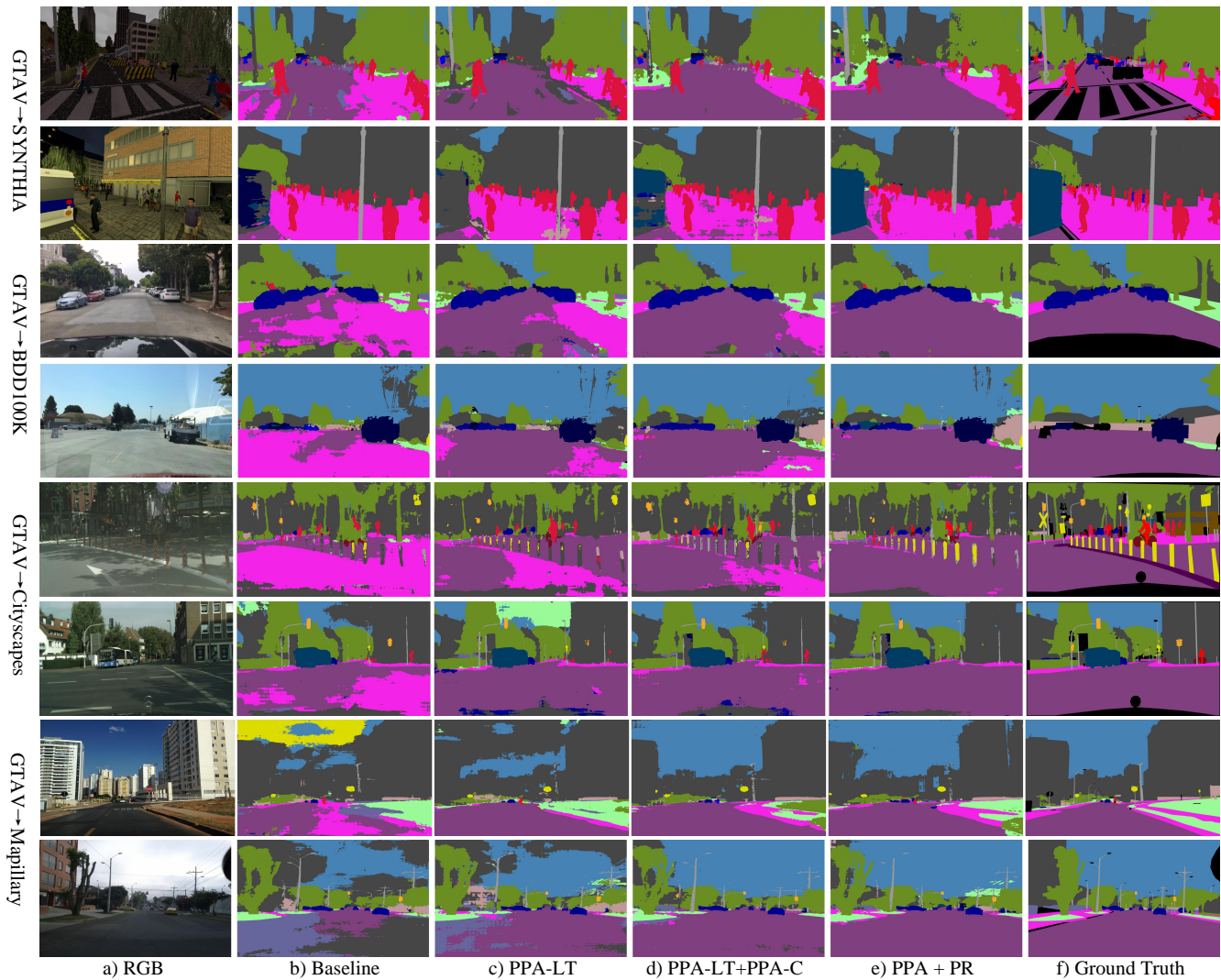}
	\caption{Qualitative examples using ResNet-50 backbone on the task of GTAV generalizing to \{B, M, C, S\}.}
	\label{visualization}
\end{figure*}

\begin{figure*}[t]
\centering
\includegraphics[width=0.98\textwidth]{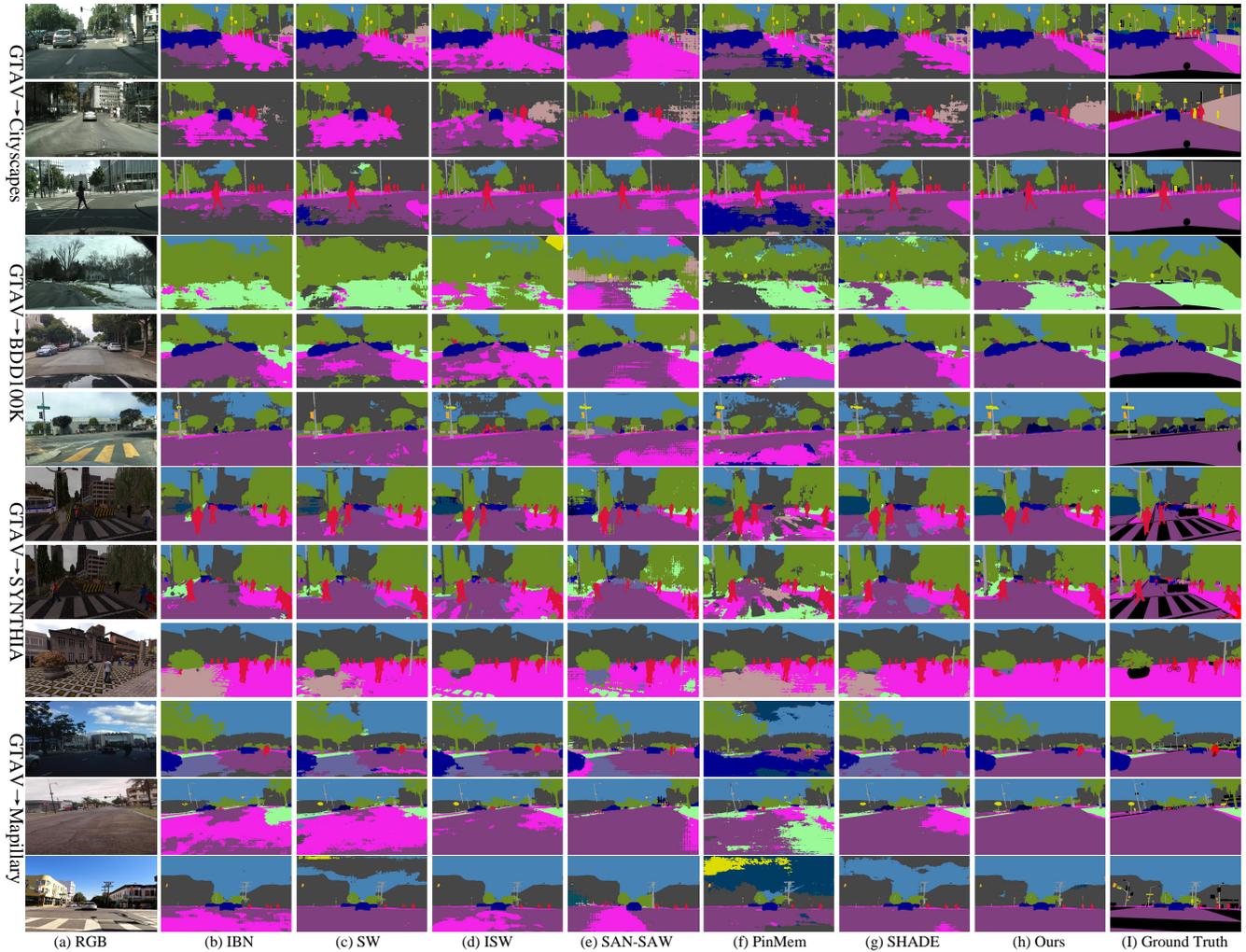}
\caption{Qualitative comparison with other SOTA methods using ResNet-50 backbone on the task of G $\rightarrow$ \{B, M, C, S\}.}
\label{SOTAvisualization}
\end{figure*}

\subsection{Ablation studies}
\subsubsection{The effect of proposed components}
Ablation studies on the proposed components including PPA (prototypical progressive alignment) and PR (prototypical reweighting) are conducted. To clearly understand the effect of each basic visual factor, the PPA is divided into PPA-LT (Prototypical progressive alignment using local texture) and PPA-C (Prototypical progressive alignment using color). As shown in Table \ref{table:ABC}, the performance using a single module has at least a 3.2\% gain in terms of mIoU compared to the baseline model \cite{choi2021robustnet}. The model with the PR learning strategy has a performance gain of 3.6\% in mIoU compared to the baseline, showing that this module can reduce negative transfer by reducing the effect of the source outliers. Furthermore, it is shown that there is a 4.4\% performance gain for all the performances using the two proposed components compared to the baseline model, from 35.1\% to 39.5\%. By further introducing the third component, the final performance can be improved to 40.1\%. The clear increase in performance shows the effectiveness of the proposed components.

\begin{table}[t]
    \centering
    \tabcolsep=0.08cm
    \caption{Ablation studies on the proposed components including PPA-LT, PPA-C, and PR. The experiments are conducted using ResNet-50 on the task of G generalizing to other datasets.}

    \small
    \begin{tabular}{l|ccc|ccccc}
         \bottomrule
         Method & PPA-LT & PPA-C & PR & $\rightarrow$C &$\rightarrow$B&$\rightarrow$M & $\rightarrow$S & Avg \\
	   \cline{1-9}
        Baseline &&  && 36.6 & 35.2&{40.3}& 28.3 & 35.1 \\
        PPA-LT &\checkmark& & &40.7 &{39.2}&{43.5}&31.7 & 38.8 \\
		PPA-C &  & \checkmark& & 40.4 & {39.2} & 42.3 & 31.3 & 38.3\\
	  PR & & &\checkmark & 41.6 & \textbf{40.1} & 41.5 & 31.5 & 38.7\\
		PPA-LT + PPA-C & \checkmark& \checkmark &  & 40.8 & \uline{40.0} & \textbf{45.3} & {31.9} & \uline{39.5} \\
	   PPA-LT + PR & \checkmark& & \checkmark &\textbf{42.8} &38.6 &42.5 & \textbf{34.2} & \uline{39.5} \\
 		PPA-C + PR & & \checkmark& \checkmark & \textbf{42.8}& 37.8 & 43.0 & \textbf{34.2} &\uline{39.5} \\ 
		ALL & \checkmark & \checkmark&  \checkmark &\uline{42.7}& {39.7} &\uline{44.5} &\uline{33.6} &\textbf{40.1}\\
         \toprule
    \end{tabular}
\label{table:ABC}
\end{table}

\subsubsection{Comparison with naive prototype and alignment methods}
Also, we provide the performance comparison with naive alignment methods, including SPA and MPA shown in Fig. \ref{Fig:PROTOCOMP}. Meanwhile, such methods are implemented using the naive prototype (NP), the OTP, and the VTP for comparison. For MPA, the shallow feature alignment is conducted in the Stage 1 feature. Four observations can be seen in Table \ref{table:othermix}. Firstly, the MPA method is slightly better than the SPA. Secondly, the text prototype alignment including OTP and VTP shows better generalizability than the NP, achieving an improvement of at least 0.9\% in both alignment forms. The reason is the naive prototype treats all features as the same causing negative transfer, which goes against with the viewpoint that different source features have different adaptation difficulties pointed out by some previous works \cite{yu2021dast, wang2019characterizing}. Instead, our method adopts more generalized text prototypes, reducing the effect of outliers and thus mitigating negative transfer. Thirdly, the model using VTP is slightly worse than the model utilizing OTP. A reasonable explanation is that the style information of VTP affects performance in the deep future alignment because the deep feature is expected to carry more category information \cite{gatys2016image, zhang2018fully}. Fourthly, our proposed PPA module outperforms SPA and MPA methods by a significant margin. This demonstrates the effectiveness of the text prototype and progressive alignment. Furthermore, the time complexity is evaluated and shown in Table \ref{table:Timecomp}. $\downarrow$ denotes that the lower value is better. Since our method mapped the dimension of features of the last layer as 512 instead of 256 to match the dimension of the CLIP text feature, it can be seen that the a slight increase in the model parameter and GFLOPs. Note that as the CLIP text model is not used during the evaluation stage, its parameters are not included. During the training stage, compared to the methods using NP, our method has a similar iteration time and training memory but has an obvious performance gain, showing the competitive efficiency of our method in text prototype generation. 

\begin{table}[t]
    \centering
    \tabcolsep=0.2cm
    \caption{Performance comparison on naive alignment methods, which includes SPA and MPA. These methods are conducted using the naive prototype (NP), the OTP, and the VTP.}
    \small
    \begin{tabular}{l|ccccc}
         \bottomrule
         Method & $\rightarrow$C & $\rightarrow$B & $\rightarrow$M & $\rightarrow$S & Avg  \\
	   \cline{1-6}
        Baseline & 36.6 & 35.2 & 40.3 & 28.3 & 35.1  \\
	   SPA-NP & 38.4&36.4& 39.5 & 30.6 & 36.2 \\
	   MPA-NP & 39.4 & 35.6 &  39.0& 30.9 &36.2  \\
      SPA-OTP & 40.5& 37.2 &39.9& 33.2 & 37.7   \\
      MPA-OTP  & \textbf{40.9} & {38.4} & {40.5} & \textbf{31.9} & \uline{37.9}   \\
	SPA-VTP & 39.9 & 38.1 & 40.1 & 30.1  & 37.1  \\
	MPA-VTP &  40.3 & \uline{38.5} & \uline{41.0} & 30.3 & 37.5 \\
	PPA (Ours) & \uline{40.8} & \textbf{40.0} & \textbf{45.3} & \textbf{31.9} & \textbf{39.5}  \\
         \toprule
    \end{tabular}
\label{table:othermix}
\end{table}
\begin{comment}

\end{comment}
\begin{table}[t]
	\centering
	\tabcolsep=0.1cm
	\caption{Time complexity comparison on naive alignment methods, which includes SPA and MPA. These methods are conducted using the naive prototype (NP), the OTP, and the VTP. Param, $t_i$, $t_p$, and Mem represent model parameter, iteration time, prototype generation time, and training memory, respectively.}
	\small
	\begin{tabular}{l|ccccc}
		\bottomrule
		Method &  GFLOPs $\downarrow$ & Param (M) $\downarrow$ & $t_i$ (s) $\downarrow$ & $t_p$ (s)  $\downarrow$ & Mem (M) $\downarrow$ \\
		\cline{1-6}
		Baseline &  555.5 & 45.08& 0.254& 0.000 & 8484 \\
		SPA-NP & 633.5 &45.68 &0.298 &0.040 & 10196 \\
		MPA-NP &  633.5& 45.68& 0.325& 0.068&  10152 \\
		SPA-OTP & 633.5& 45.68& 0.279& 0.024& 10300   \\
		MPA-OTP  &  633.5& 45.68& 0.321& 0.065& 11580  \\
		SPA-VTP &  633.5& 45.68& 0.281& 0.027 & 10300 \\
		MPA-VTP & 633.5 &45.68& 0.318&  0.063 &11582 \\
		PPA (Ours) & 633.5 &45.68 &0.315 & 0.060&11580  \\
		\toprule
	\end{tabular}
	\label{table:Timecomp}
\end{table}

\subsubsection{Qualitative comparison}
Fig. \ref{visualization} shows some results from different models of our proposed method. Evidently, the accuracy of results gradually improves by using more proposed components.
\par Fig. \ref{SOTAvisualization} shows qualitative comparison with other SOTA methods including IBN \cite{pan2018two}, SW \cite{pan2019switchable}, ISW\cite{choi2021robustnet}, SAN-SAW \cite{peng2022semantic}, PinMem \cite{kim2022pin}, SHADE\cite{zhao2022style}. Obviously, our PPAR method achieves better segmentation results.
\begin{table}[t]
	\centering
	\tabcolsep=0.2cm
	\caption{Ablation studies on the selection of alignment layer in shallow feature alignment. DA and SA represent deep and shallow alignment, respectively. SA1-3 denotes the shallow alignment at the feature of Stages 1-3, respectively.}
	
	\small
	\begin{tabular}{l|ccccc}
		\bottomrule
		Method & $\rightarrow$C & $\rightarrow$B & $\rightarrow$M & $\rightarrow$S & Avg\\
		\cline{1-6}
		Baseline & 36.6 & 35.2 & 40.3 & 28.3 & 35.1 \\
		DA + SA1 & \textbf{40.8} & \textbf{40.0} & \textbf{45.3} & \uline{31.9} & \textbf{39.5} \\
		DA + SA2 & \uline{40.4} & \uline{38.9} & 41.8 & \textbf{32.9} & \uline{38.5} \\
		DA + SA3 & 40.3 & 38.4 & 42.2 & 31.7 & 38.2 \\
		DA + SA1 \& 2  & 39.8 & 38.4 & 40.0 & 33.4 & 37.9 \\
		DA + SA1 \& 3  & 40.0 & 37.3 & 40.3 & 31.7 & 37.3 \\
		%Stages 1 \& 2 \& 3 & 39.9 &37.9 &\uline{43.8} &30.0 &37.9 \\
		\toprule
	\end{tabular}
	\label{table:ASALS}
\end{table}

\subsubsection{Alignment Layer validation}
As previously mentioned, the PPA strategy consists of shallow feature alignment and deep feature alignment. For deep feature alignment, as pointed out by the existing work \cite{zhang2018fully}, deep feature carries more semantic information rather than basic visual information for better classification. Hence, to follow this observation, the OTP is employed to align the deep feature. Then, the concern turns into the selection of the alignment layer for shallow feature alignment. To verify the effect of PPA using the feature extracted in different layers, ablation studies of different shallow features are conducted. Specifically, the shallow feature alignment is conducted in Stages 1-3 and the deep feature alignment is conducted in Stage 4. The ablation results are shown in Table \ref{table:ASALS}. Obeying the easy-to-difficult alignment principle, for the single-stage setting ($2-4^{th}$ rows in Table \ref{table:ASALS}), PPA-LT and PPA-C are adopted. For the two-stage setting ($5-6^{th}$ rows in Table \ref{table:ASALS}), the first stage adopts the above setting (i.e., PPA-LT + PPA-C), and the second stage adopts PPA-LT.

\par It can be seen that the performance achieves the best when the shallow feature alignment is conducted using the Stage 1 feature. For the single-stage setting, the performance gradually decreases from Stage 1 to Stage 3, which shows that preserving basic visual information in the shallow feature is better than in the deep feature. For the two-stage setting, the performance is slightly worse than the model using the single-stage setting. These results show the significance of preserving basic visual information in the shallow feature.
\begin{table}[t]
	\centering
	\tabcolsep=0.2cm
	\caption{Method applicability validation.}
	\small
	\begin{tabular}{l|ccccc}
		\bottomrule
		Method & $\rightarrow$C & $\rightarrow$B & $\rightarrow$M & $\rightarrow$S & Avg  \\
		\cline{1-6}
		SW \cite{pan2019switchable} &29.9 &27.5 &29.7 &27.6 &28.7 \\
		SW + Ours & \textbf{40.7}& \textbf{34.7}& \textbf{38.6}& \textbf{30.7} &\textbf{36.2} \\
		\hline
		IBN \cite{pan2018two} & 33.9 &32.3 &37.8& 27.9 &33.0 \\
		IBN + Ours &\textbf{40.8}& \textbf{36.0} &\textbf{41.3} &\textbf{30.4} &\textbf{37.1} \\
		\hline
		ISW  \cite{choi2021robustnet}& 36.6 &35.2 &40.3 &28.3& 35.1 \\ 
		ISW + Ours & \textbf{42.7} & \textbf{39.7} & \textbf{44.5} & \textbf{33.6} & \textbf{40.1} \\
		\hline
		SHADE \cite{zhao2022style} &\textbf{43.5}& 40.3 &43.0 &31.2 &39.5  \\
		SHADE + Ours & 42.5 & \textbf{41.7} & \textbf{43.8} & \textbf{34.1} &\textbf{40.5}  \\
		\toprule
	\end{tabular}
	\label{table:mgmix}
\end{table}

\begin{figure}[t]
	\centering
	\includegraphics[width=0.48\textwidth]{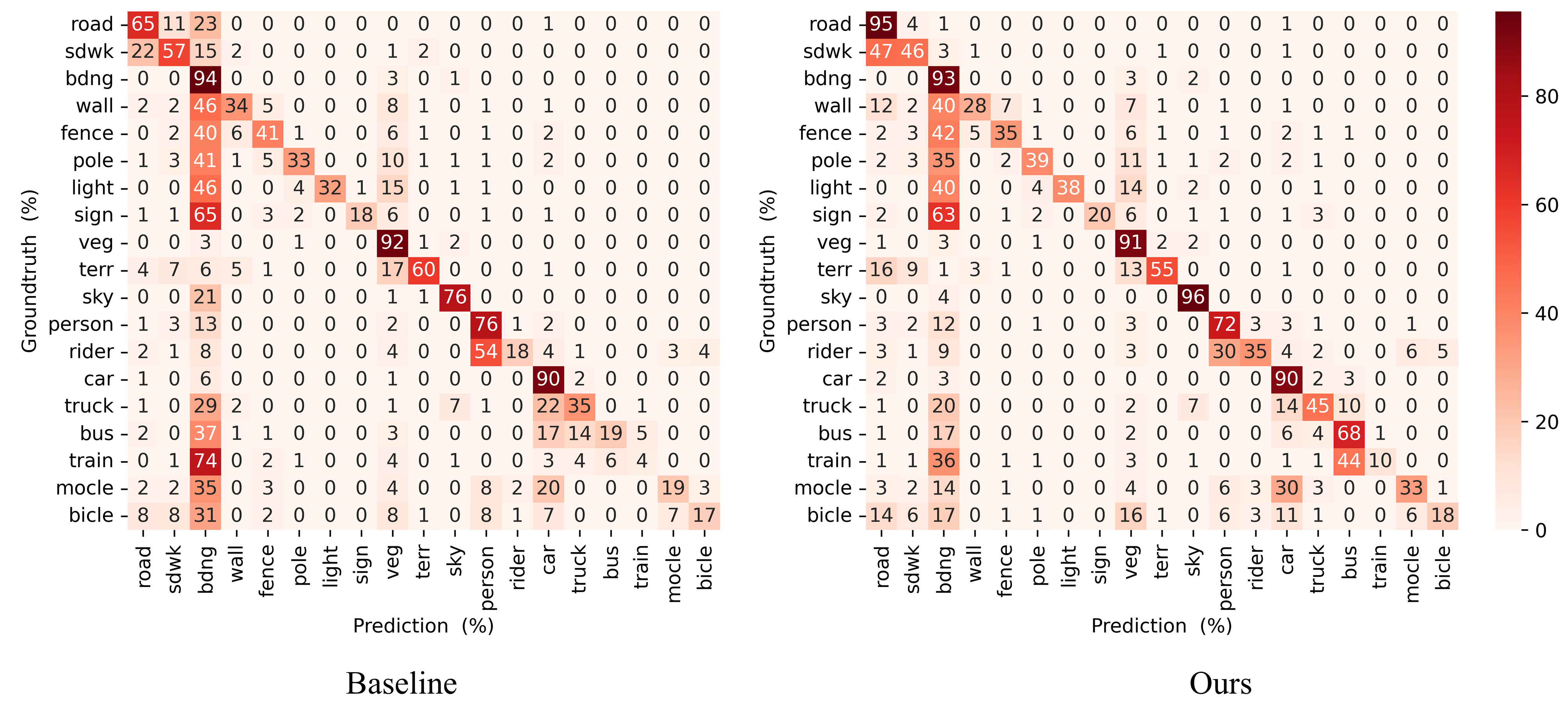}
		\caption{Confusion matrix comparison between the baseline and ours in the setting of GT5V generalizing to Cityscapes. Zoom in to see clearly.}
	\label{fig:confumatrix}
\end{figure}
\subsubsection{Method generalization validation}
To verify the applicability of our method, we plug our method into other methods. The experiments are conducted using ResNet-50 on the task of GTAV generalizing to other datasets. As depicted in Table \ref{table:mgmix}, when integrating our method into SW \cite{pan2019switchable}, IBN \cite{pan2018two}, ISW \cite{choi2021robustnet}, and SAHDE \cite{zhao2022style}, the performances are respectively enhanced by 7.5\%, 4.1\%, 5.0\%, and 1.0\% in average mIoU, demonstrating the applicability of our method.
\subsubsection{Confusion matrix comparison}

We show the confusion matrix comparison between the baseline and our method. A high diagonal value shows high accuracy and a high non-diagonal value indicates worse prediction. As shown in Fig. \ref{fig:confumatrix}, our method has 11 classes performance better than the baseline, especially the ``road", ``sky", ``rider", and ``bus" classes.  Meanwhile,   our method also has lower values on non-diagonal lines generally. A concrete example is the third column of confusion matrixes, many samples of different classes are incorrectly segmented as the building class. Our method has 15 lower values in this column. These analyses show that our method has better accuracy and reduces class confusion.

\subsubsection{T-SNE visualization}
\begin{figure}[t]
	\centering
	\includegraphics[width=0.48\textwidth]{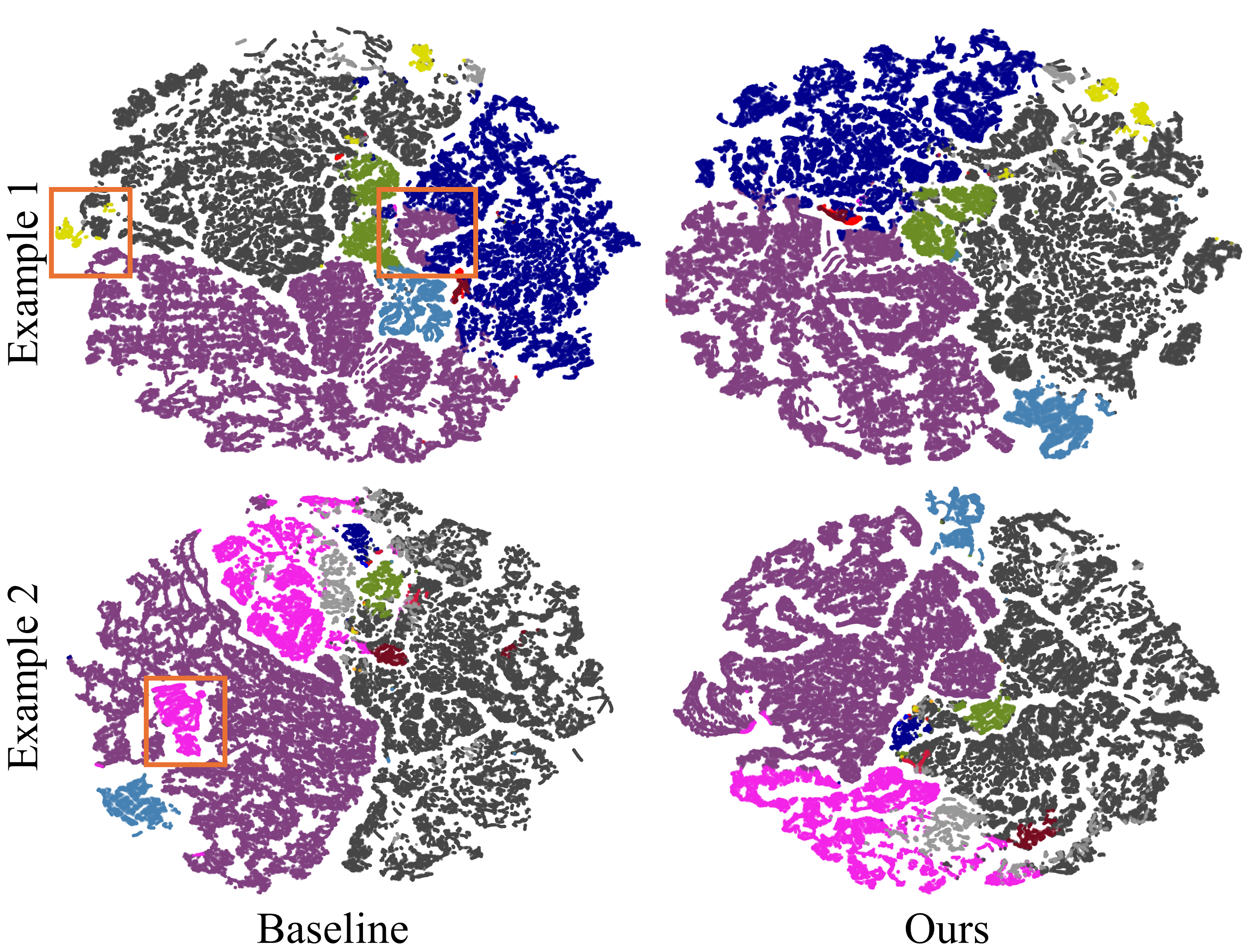}
	\caption{T-SNE visualization comparison between the baseline and ours in the setting of GT5V generalizing to Cityscapes. Orange rectangles show the classes far from the main clusters.}
	\label{fig:TSNEcompa}
\end{figure}

T-SNE \cite{van2008visualizing} is a widely-used visualization method mapping high-dimensional data to two dimensions. To thoroughly comprehend the advantage of our method, it is valuable to analyze the discrepancies in T-SNE graphs between the baseline and our method. Fig. \ref{fig:TSNEcompa} illustrates two concrete examples. As shown in the first row of Fig. \ref{fig:TSNEcompa}, part of the ``traffic light" (depicted in yellow) and ``road" (depicted in purple) classes appear disentangled from the main clusters, which potentially results in class confusion. Similarly, part of the ``sidewalk" is isolated by ``road", shown in the second row of Fig. \ref{fig:TSNEcompa}. Conversely, our method alleviates this situation, thereby reducing the class confusion problem.

\section{Limitations and Future Research}
Although our method achieves superior performance, there are some limitations. First, our method is limited by the quality of the CLIP model. Specifically,  as the CLIP model is learned from many large-scale and diverse datasets, reducing the distribution gap between class-wise features and text prototypes can implicitly learn knowledge from large-scale datasets, which can be seen as an implicit knowledge-distillation process. Considering that text prototypes are not learnable in our module, it is foreseeable that the class-wise feature will gradually approximate the text prototype. Thus, it is reasonable that better text prototypes can be generated by the model trained using more large-scale datasets or novel training strategies, causing better generalizability of our method. Second, recently some DG models like Rein \cite{wei2024stronger} employing visual foundation models show their effectiveness, while only language foundation model is utilized in our method. Combining both language and vision foundation models to improve generalization is an interesting and novel idea. Furthermore, our method focuses on unseen domain generalization without considering a novel unseen category. It is still an unexplored field to cogitate unseen domain generalization and unseen class recognition simultaneously, which is closer to the real-world situation.
\section{Conclusion}
In this paper, to improve the effectiveness of prototypical alignment and alleviate negative transfer, we propose a novel method for generalizable semantic segmentation, called Prototypical Progressive Alignment and Reweighting (PPAR) depending on the strong generalized representation of the CLIP model. First, we define the VTP and the OTP extracted by the CLIP model, which provides a more generalizable representation for prototypical alignment. Then, to fully utilize the VTP, we propose a prototypical progressive alignment strategy by an easy-to-difficult alignment form to reduce domain-variant information progressively instead of directly. Finally, we propose a prototypical reweighting learning strategy to alleviate negative transfer by estimating the importance of the source data and correcting its learning weight. The extensive experiments show that our model achieves state-of-the-art results, proving the effectiveness of the proposed PPAR method.

\bibliography{egbib} %.bib文件名字
\bibliographystyle{IEEEtran} %.bst模板

\end{document}